\documentclass{article}

\usepackage{arxiv}

\usepackage[utf8]{inputenc} % allow utf-8 input
\usepackage[T1]{fontenc}    % use 8-bit T1 fonts
\usepackage{hyperref}       % hyperlinks
\usepackage{url}            % simple URL typesetting
\usepackage{booktabs}       % professional-quality tables
\usepackage{amsfonts}       % blackboard math symbols
\usepackage{nicefrac}       % compact symbols for 1/2, etc.
\usepackage{microtype}      % microtypography
\usepackage{lipsum}
\usepackage{array}
\usepackage{amsmath}
\usepackage{graphicx}
\usepackage{subfigure}
\usepackage{tabularx}
\usepackage{varwidth}
\usepackage[linesnumbered,lined,boxed,commentsnumbered, ruled]{algorithm2e}
\usepackage{url}
\usepackage{multirow}
\usepackage{bm}
\usepackage{color}
\usepackage[dvipsnames]{xcolor}

\title{Drug-Drug Interaction Prediction with Wasserstein Adversarial Autoencoder-based Knowledge Graph Embeddings}

\author{
  Yuanfei Dai \\
  College of Mathematics and Computer Sciences\\
  Fuzhou University\\
  Fujian, China\\
  \texttt{daiyuanfly@gamil.com} \\
   \And
 Chenaho Guo \\
  College of Mathematics and Computer Sciences\\
  Fuzhou University\\
  Fujian, China \\
  \texttt{guochenhao@gmail.com} \\
    \And
 Wenzhong Guo \\
  College of Mathematics and Computer Sciences\\
  Fuzhou University\\
  Fujian, China \\
  \texttt{guowenzhong@fzu.edu.cn} \\
    \And
 Carsten Eickhoff \\
  Center for Biomedical Informatics\\
  Brown University\\
  Providence, RI, USA \\
  \texttt{carsten@brown.edu} \\
}

\begin{document}
\maketitle

\begin{abstract}
Interaction between pharmacological agents can trigger unexpected adverse events. Capturing richer and more comprehensive information about drug-drug interactions (DDI) is one of the key tasks in public health and drug development. Recently, several knowledge graph embedding approaches have received increasing attention in the DDI domain due to their capability of projecting drugs and interactions into a low-dimensional feature space for predicting links and classifying triplets. However, existing methods only apply a uniformly random mode to construct negative samples. As a consequence, these samples are often too simplistic to train an effective model. In this paper, we propose a new knowledge graph embedding framework by introducing adversarial autoencoders (AAE) based on Wasserstein distances and Gumbel-Softmax relaxation for drug-drug interactions tasks. In our framework, the autoencoder is employed to generate high-quality negative samples and the hidden vector of the autoencoder is regarded as a plausible drug candidate. Afterwards, the discriminator learns the embeddings of drugs and interactions based on both positive and negative triplets. Meanwhile, in order to solve vanishing gradient problems on the discrete representation--an inherent flaw in traditional generative models--we utilize the Gumbel-Softmax relaxation and the Wasserstein distance to train the embedding model steadily. We empirically evaluate our method on two tasks, link prediction and DDI classification. The experimental results show that our framework can attain significant improvements and noticeably outperform competitive baselines.
\end{abstract}

% keywords can be removed

\section{Introduction}
\label{section1}
For optimal therapeutic effect, it is often necessary to take advantage of drug combinations. However, the intended efficacy of a drug may be changed substantially when co-administered alongside another agent. Formally, drug-drug interactions (DDI) are pharmacological interactions between drug ingredients which can alter the function of drugs, cause adverse drug reactions (ADR) and even medical malpractice~\cite{cheng2014machine}. While ideally we would like to discover all possible interactions between drugs during clinical trial, some unrecognized interactions may only be revealed after the drugs are approved for clinical use. ADRs cause roughly 100,000 fatalities~\cite{giacomini2007good} and 74,000 emergency room visits in the United States, annually \cite{percha2013informatics}. For instance, acetylsalicylic acid (ASA), also known as aspirin, is a commonly used drug for the treatment of fever and pain, which has both anti-inflammatory and antipyretic effects. However, when ASA is combined with 1-benzylimidazole, the risk or severity of hypertension can be increased. To alleviate these risks and improve quality of care, large-scale and reliable DDI prediction becomes a key task in clinical practice.

To date, various DDI prediction approaches have been proposed to solve this issue. Examples from the fields of pharmacogenomics and pharmacology include~\cite{jia2009mechanisms, palleria2013pharmacokinetic}. However, these methods can only handle a limited range of DDI cases because of their dependency on clinical and laboratory data. Besides, this kind of method requires many characteristics such as molecular structures, pharmacology, indications, \textit{etc.} of each drug. For this reason, knowledge graph-based computational prediction approaches which do not rely on these expensive, labor intensive features have received ever-increasing attention due to their capability of enabling automatic, fast assessments of possible drug–drug interactions.

\begin{figure}
\centering
\includegraphics[width=11cm]{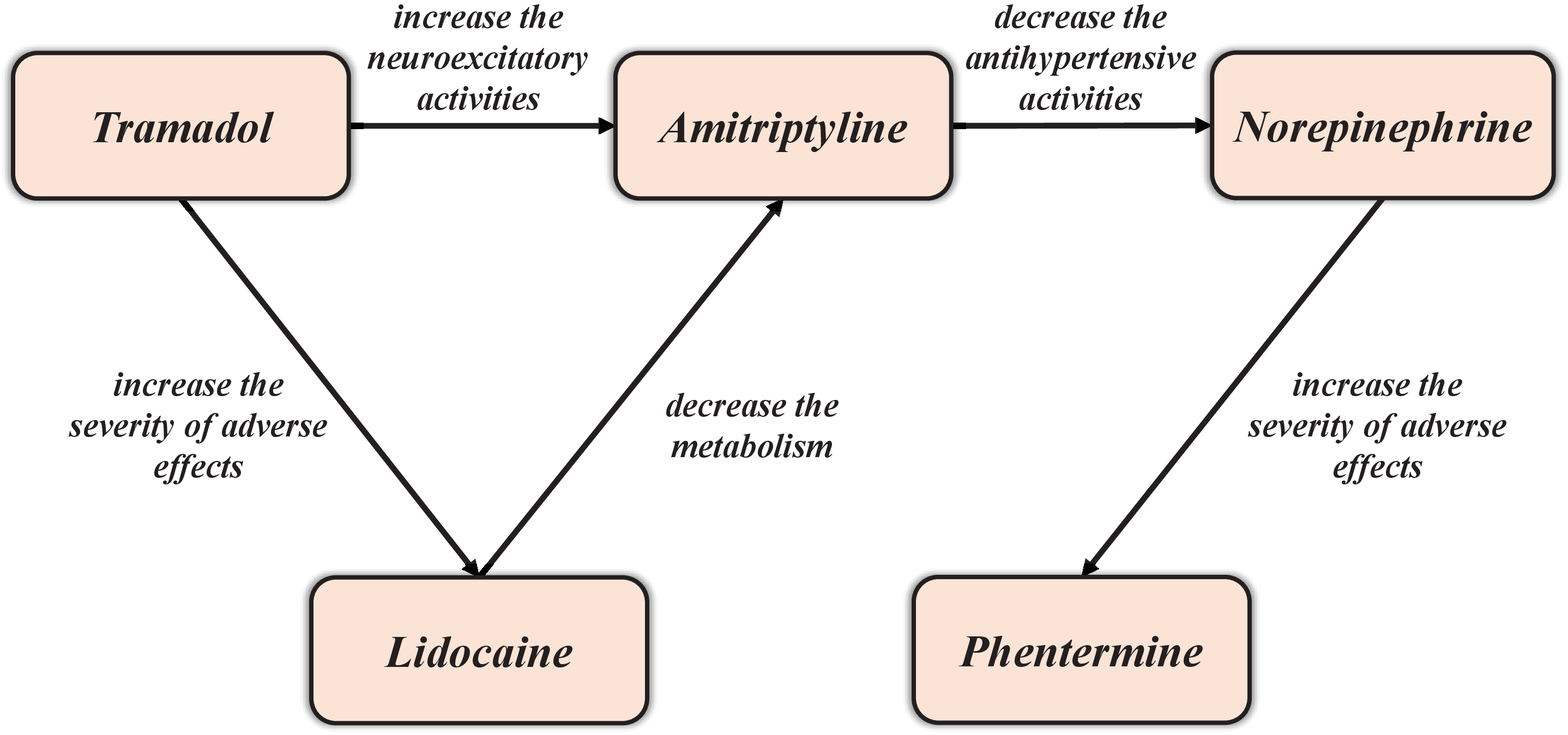}
\caption{A simple instance of a DDI knowledge graph.}
\label{Figure1}
\end{figure}

DDI data can be represented as a knowledge graph (KG) in which nodes indicate entities and edges denote relations. A typical DDI knowledge graph is constructed with a series of triplet facts $(h, r, t)$ in which $h$ and $t$ represent head and tail drugs respectively, and $r$ indicates the interaction between $h$ and $t$. Accordingly, the DDI prediction problem can be posed as a link prediction task via knowledge graph embedding, which aims to embed each entity and relation to a low-dimensional feature space for knowledge fusion and more efficient computing. Figure~\ref{Figure1} shows an example of a DDI knowledge graph.

Over the past years, several machine learning and deep learning approaches have been proposed to embed DDI knowledge graphs for predicting unknown DDIs~\cite{ma2019genn, ryu2018deep, zitnik2018modeling}. Firstly, training a KG embedding model requires negative samples and there are no confirmed negatives in the original DDI datasets. In order to conveniently generate enough negative samples to train the model, Bordes et al.~\cite{transE2013} introduce a local closed-world assumption (LCWA) into the knowledge graph. Under the LCWA, all statements that exist in the knowledge graph are assumed correct. Conversely, any statements that do not exist are false. This assumption is conducive to the generation of negative samples, as we only need to construct triplets which are not contained in the original knowledge graph to be considered as fake samples. Most of existing embedding models have been generating negative triplets via a uniform negative sampling strategy ~\cite{transE2013,transA2016,kazemi2018simple}. This sampling randomly replaces the head or tail entity in a positive triplet with a different one from the entity collection, where all entities share the same sampling weights.  Trouillon et al.~\cite{complex2016} evaluated the performance impact of the number of negative triplets constructed per positive training sample. The results revealed that generating more negatives can, up to a saturation threshold, yield better performance. However, this general sampling method usually adds only limited benefit to the robustness and effectiveness of the derived embedding model and may even delay model convergence as noted by Schroff et al.~\cite{schroff2015facenet} and Hermans et al.~\cite{hermans2017defense}. Thus, we utilize adversarial learning to generate more plausible negative triplets for improving the performance of knowledge graph representation learning. For instance, if we want to replace the head drug in an observed triplet (Tramadol, increase neuroexcitatory activities, Amitriptyline) to construct a negative triplet between the two candidates ``Ibuprofen'' and ``Nexium'', ``Ibuprofen'' makes for a more deceptive replacement due to its pharmacological similarity to ``Tamadol''. Afterwards, this more plausible triplet can force the KG embedding model to improve performance to distinguish its authenticity. Nevertheless, it is also likely to choose other irrelevant drugs that would make it easy for the embedding model to distinguish and does not encourage it to improve (\textit{e.g.}, ``Nexium'') if the head drug is replaced by the above random sampling mode. Our proposed method represents all drugs in a unified feature space, and then selects a suitable drug as a substitute to generate a deceptive negative sample according to the spatial position and distance between the drugs, thereby further improving the performance of the model.

Unfortunately, adversarial learning methods such as generative adversarial networks (GAN) have not yielded satisfying results for natural language processing tasks as the standard GAN is limited to the continuous real number space, \textit{i.e.}, continuous data, but cannot directly operate on discrete data such as words. To overcome this deficiency, recent research provides a number of feasible approaches by applying policy gradients, a class of policy-based reinforcement learning algorithms, to replace the traditional back propagation~\cite{wang2018incorporating, kbgan2018}.

Although these reinforcement learning (RL) approaches have been proven effective, high variance gradient estimates make models require vast amounts of computational resources while their complex hyperparameters increase instability of the already difficult-to-train GANs. In this work, we propose a new method which introduces Gumbel-Softmax relaxation~\cite{jang2016categorical, maddison2016concrete} and adversarial autoencoders (AAE) based on Wasserstein distances for training DDI embedding models steadily on discrete data. In contrast to complicated RL mechanisms, the Gumbel-Softmax relaxation can efficiently simplify our model and allow for a fast iterative adversarial learning framework without intensive RL heuristics for accelerating the convergence of the entire model. Compared to GANs, AAEs can control the manner in which the generator constructs negative samples, making their outputs resemble real data more closely. Furthermore, we use the Wasserstein distance as an advanced metric to replace the original Jensen-Shannon (JS) divergence in the traditional adversarial learning framework. 

To this end, we first construct an autoencoder where the latent code vector $\bm{z}$ (\textit{i.e.}, its hidden units) is trained to generate more plausible entities (drugs) as negative samples. Since the entity we intend to generate is a one-hot vector and this type of discrete data is not differentiable in the training process, a Gumbel-Softmax relaxation and the Wasserstein distance are employed to handle the issue of vanishing gradients on discrete data without policy gradient mechanisms. Then, negative and positive triplets are jointly fed into the discriminator to obtain the embeddings which are regarded as the final representation of the KG. Our novel contributions in this paper are summarized as follows:
\begin{itemize}
	\item We present a new approach to solve the prediction of drug-drug interactions and their side-effects. Compared with clinical trials or traditional machine learning based methods, our approach does not require numerous manual features to yield better performance.
	
	\item Technically, to the best of our knowledge, we are the first to introduce adversarial autoencoders to knowledge graph representation learning. The latent vector of the autoencoder is capable of generating more reasonable negative samples and the discriminator utilizes these negative and positive triplets to train the KG embedding model.
	
	\item Different from traditional adversarial learning for KG embedding which requires intensive RL heuristics, we apply Gumbel-Softmax relaxation and Wasserstein distance to resolve the problem of vanishing gradients on discrete data and accelerate the convergence of the KG embedding model.
	
	\item We evaluate the performance of the proposed model on link prediction and triple classification tasks. The experimental results show that our model outperforms existing KG embedding models.
\end{itemize}

The remainder of this article is organized as follows. Section~\ref{section2} introduces related work on drug-drug interaction detection and prediction and several representative knowledge graph embedding models. In Section~\ref{section3}, we illustrate the overall framework and training procedure of the proposed adversarial learning model in detail. Section~\ref{section4} delineates benchmark datasets, parameter initialization settings and experimental details. Section~\ref{section5} provides a side-by-side qualitative comparison and discussion between our results and those obtained via existing methods. Finally, concluding remarks are discussed in Section~\ref{section6}.
%%%%%%%%%%%%%%%%%%%%%%%%%%%%%%%%%%%%%%%%%%%%%%%%%%%%%%%%%%%%%%%%%%%%%%%%%%%%%%%%%%%%%
\section{Related Work}
\label{section2}

In this section, we introduce current research on drug-drug interaction detection and prediction. Additionally, we give a brief overview of several prominent existing knowledge graph embedding methods.

\subsection{Drug-drug interaction detection and prediction}
DDI prediction is a key task in pharmacology. Many existing  studies that obtain results on specific types of interactions depend on \textit{in vivo} and \textit{in vitro} experiments. Krishna et al.~\cite{krishna2009pharmacokinetics} constructed a crossover study to evaluate the effects of gastric pH values on the absorption of posaconazole. The results displayed that both dissolution and absorption of posaconazole would be decreased under increased gastric pH conditions (\textit{e.g.}, induced via co-administration with proton pump inhibitor drugs, such as esomeprazole or omeprazole). To reveal the clinical effect of omeprazole on the inhibition of platelet clopidogrel, Ho et al.~\cite{ho2009risk} conducted a retrospective cohort study of 8,205 patients, finding that the risk of adverse outcomes would be increased when clopidogrel is accompanied by proton pump inhibitors. Menon et al.~\cite{menon2015drug} evaluated DDIs between the 3D regimen of direct-acting antiviral agents for the treatment of chronic hepatitis C virus infection (such as ombitasvir, Paritaprevir, and dasabuvir) and various common medications via 13 experiments. Although the above works yield detailed comparative results, they do not scale well due to laboratory requirements. With the advancement of computing methods and resources, researchers moved their attention towards large-scale structured databases and machine learning based approaches to solve this problem.

Several studies have proposed automatic DDI discovery schemes. For instance, Cheng and Zhao~\cite{cheng2014machine} introduced phenotypic, therapeutic, genomic and chemical structural similarity as drug features and employed five machine learning approaches, including $k$-nearest neighbors, na\"{i}ve Bayes, logistic regression, decision trees and support vector machines, to predict the authenticity of DDIs. Li et al.~\cite{li2015large} constructed a Bayesian network for forecasting the combined effect of drugs by incorporating drug molecular and pharmacological phenotypes. Lately, to further enhance the performance of the DDI prediction model, many semantic and topological measures between drugs are utilized as input features for discovering potential DDIs ~\cite{kastrin2018predicting}. Mu{\~n}oz et al.~\cite{munoz2019facilitating} utilized knowledge graphs as a convenient uniform representation to integrate multiple forms of heterogeneous data, so that the data can be represented by a unified feature description. However, these feature-based approaches not only rely heavily on the quality of hand-crafted features, but also suffer from issues of data incompleteness and sparsity.

Knowledge graph embedding has received increasing attention due to its strong capability of overcoming data incompleteness and sparsity problems. KG embedding methods have been demonstrated to offer competitive performance in DDI prediction tasks. Among them, Abdelaziz et al.~\cite{abdelaziz2017large} proposed a large-scale framework for DDI prediction, called Tiresias. It first integrated various drug-related variables as a DDI knowledge graph, then leveraged this KG to compute several similarity measures between all the drugs and predicted potential DDIs via a logistic regression classifier. Celebi et al.~\cite{celebi2018evaluation} applied several classic knowledge graph embedding algorithms such as TransE and TransD to extract feature vectors in order to predict potential interactions between drugs. Ma et al.~\cite{ma2018drug} used multi-view graph autoencoders to integrate multiple types of drug-related information, and added an attention mechanism to calculate the corresponding weights of each view for better interpretability. Zitnik et al.\cite{zitnik2018modeling} developed a graph convolutional neural network in which an end-to-end model was built for multi-relational link prediction on a multi-modal graph. Karim et al.~\cite{karim2019drug} combined ComplEx (a traditional KG embedding method) with a convolutional-LSTM network to further refine model performance.

In the following, we will discuss a representative range of knowledge graph embedding techniques in greater detail.

\subsection{Existing knowledge graph embedding approaches}
There has been an increasing amount of literature on knowledge graph embedding to represent both entities and relations in a low-dimensional continuous feature space~\cite{wang2017knowledge, ji2020survey}. We have broadly categorised these existing embedding methods into three categories: \emph{translation-based methods}, \emph{tensor factorization-based methods} and \emph{neural network-based methods}.

\subsubsection{Translation-based embedding methods}
Mikolov et al.~\cite{mikolov2013distributed} proposed translation invariance in the word embedding algorithm \emph{word2vec} that allows words with similar connotation to have similar representations. Following this principle, Bordes et al.~\cite{transE2013} proposed the \textbf{TransE} knowledge graph embedding model. \textbf{TransE} interprets relations as translation vectors between head and tail entities on the low-dimensional embedding vector space, namely $\bm{h} + \bm{r} \approx \bm{t}$. A score function is defined to measure the plausibility of each triplet fact ($h, r, t$). The score indicates the distance between $\bm{h} + \bm{r}$ and $\bm{t}$, and the function is formulated as follows:
\begin{equation}
f_{r}(h, t) = \left \| \bm{h} + \bm{r} - \bm{t} \right \| _{\ell_{1}/\ell_{2}},
\label{equ1}
\end{equation}
where $\ell_{1}$, $\ell_{2}$ are $L_{1}$-norm and $L_{2}$-norm respectively. It is worth noting that the embedding model yields a low score if a triplet $(h, r, t)$ is valid and a high score otherwise.

Although \textbf{TransE} generally delivers solid performance, it struggles to solve complex relations, such as $1-N$, $N-1$, and $N-N$. \textbf{TransH}~\cite{transH2014} was proposed to overcome this drawback by introducing relation-specific hyperplanes. \textbf{TransR} \cite{transR2015} expanded relation-specific hyperplanes to relation-specific spaces. Since then, a large number of embedding models investigated different ways to improve performance. \textbf{TransA}~\cite{transA2016} abandoned traditional Euclidean distances and adopted adaptive Mahalanobis distances as a better metric on account of their flexibility and adaptability. \textbf{TransG}~\cite{transG2016} proposed to modify the model by introducing multidimensional Gaussian distributions to replace the original conclusive numerical space and constructed a probabilistic embedding model to represent entities and relations.

\subsubsection{Tensor factorization-based embedding methods}
Tensor factorization is another effective approach to knowledge graph embedding. \textbf{RESCAL}~\cite{rescal2011} is the representative approach in this direction. Under \textbf{RESCAL}, all triplet facts in the KG are projected into a 3D binary tensor $\mathcal{X}$ to express the inherent structure, $\mathcal{X}_{ijk} = 1$ indicates that the observed triplet ($i$-th entity, $k$-th relation, $j$-th entity) exists in the graph; otherwise, $\mathcal{X}_{ijk} = 0$ refers to an unknown or non-existent triplet. Afterwards, rank-$d$ factorization is applied to obtain latent semantics in the KG. The principle that this model follows is formulated as:
\begin{equation}
\mathcal{X}_{k} \approx AR_{k}A^{T}, \  for \  k=1,2,\cdots,m,
\label{equ21}
\end{equation}
where $A \in \mathbb{R}^{n \times d}$ is a matrix which has the capability of capturing the latent semantic structure of entities and $R_{k} \in \mathbb{R}^{d \times d}$ is a matrix that models the pairwise interactions in the $k$-th relation. According to this principle, the score function $f_{r}(h,t)$ is defined as:
\begin{equation}
f_{r}(h,t) = \bm{h}^{\top}\bm{M}_{r}\bm{t},
\end{equation}
where $\bm{h}$, $\bm{t} \in \mathbb{R}^{d}$ represent embedding vectors of entities like in the above models, the matrix $\bm{M}_{r} \in \mathbb{R}^{d \times d}$ denotes the latent semantic meanings in relation $r$. To simplify the computational complexity of \textbf{RESCAL}, \textbf{DistMult}~\cite{distmult2015} restricted $\bm{M}_{r}$ to diagonal matrices, \textit{i.e.}, $\bm{M}_{r} = \mathrm{diag}(\bm{r}), \bm{r} \in \mathbb{R}^{d}$. The score function is transformed as follows:
\begin{equation}
f_{r}(h,t) = \bm{h}^{\top} \mathrm{diag}(\bm{r}) \bm{t}.
\end{equation}

The original \textbf{DistMult} model is symmetric in head and tail entities for every relation; \textbf{ComplEx}~\cite{complex2016} leveraged complex-valued embeddings to extend \textbf{DistMult} to asymmetric relations. The embeddings of entities and relations exist in the complex space $\mathcal{C}^{d}$, instead of the real space $\mathbb{R}^{d}$ in which \textbf{DistMult} embedded. The score function is modified to:
\begin{equation}
f_{r}(h,t) = \mathrm{Re} \left(\bm{h}^{\top} \mathrm{diag}(\bm{r}) \overline{\bm{t}}\right),
\end{equation}
where $\mathrm{Re}(\cdot)$ denotes the real part of a complex value, and $\overline{\bm{t}}$ represents the complex conjugate of $\bm{t}$. By using this score function, triplets that have asymmetric relations can obtain different scores depending on the sequence of entities.

\textbf{SimplE}~\cite{kazemi2018simple} proposed the inverse embedding of relations and leveraged it to calculate the average Canonical Polyadic score of $(h,r,t)$ and $(t, r^{-1}, h)$. The score function is formulated as:
\begin{equation}
    f_{r}(h, t)=\frac{1}{2}\left(\bm{h} \circ \bm{r t}+\bm{t} \circ \bm{r}^{\prime} \bm{h}\right),
\end{equation}
where $\bm{r}^{\prime}$ denotes the embedding of inversion relation and $\circ$ indicates the element-wise Hadamard product. \textbf{RotatE}~\cite{sun2018rotate} proposed a rotational model in which each relation is regarded as a rotation from source entity to target entity in complex space, as $\bm{t} = \bm{h} \circ \bm{r}$. In addition, \textbf{RotatE} also provided a self-adversarial negative sampling mode, which selects negative triplets according to the scores calculated by the current embedding model.

\subsubsection{Neural network-based embedding methods}
Deep neural networks have become popular in a multitude of fields due to their strong generalization and representation abilities. They have been widely used for knowledge graph embedding.

\textbf{ConvKB}~\cite{conv2018} was proposed to capture semantic information contained in entities and relations by incorporating convolutional neural networks (CNN). In \textbf{ConvKB}, the embedding vectors $\bm{h}$, $\bm{r}$ and $\bm{t}$ are concatenated to a matrix as an input layer, and after a convolution operation, the final output is obtained.

Inspired by adversarial learning, Minervini et al.~\cite{minervini2017adversarial} proposed an adversarial set regularization method for regularizing traditional embedding models, where an adversary samples the most plausible set of input representations. De Cao and Kipf~\cite{de2018molgan} adopted generative adversarial networks (GAN) to generate molecules with specific desired chemical properties. Moreover, Wang et al.~\cite{wang2018incorporating} and Cai et al.~\cite{kbgan2018} applied GANs to sample plausible negative training examples for KG embedding via policy gradients. They employed the generator $G(z;\theta)$ to construct negative triplets and utilized the discriminator $D(x; \phi)$ as an embedding model to distinguish artificial from real triplets.

In summary, most previous methods used random sampling or generative adversarial networks to generate negative training triplets. While GANs improved model performance, they also drastically increased computational complexity and brought instability to the training process. In this paper, we describe a new framework based on adversarial autoencoders for improving the representation ability of models by generating high quality plausible negative samples to train the discriminator. Compared with the above methods, our framework can generate more reasonable negative samples in less time, thereby improving the performance and practical usefulness of the embedding model.
%%%%%%%%%%%%%%%%%%%%%%%%%%%%%%%%%%%%%%%%%%%%%%%%%%%%%%%%%%%%%%%%%%%%%%%%%%%%%%%%%%%%%
\section{Method}
\label{section3}
A knowledge graph is a directed graph in which the nodes correspond to entities and the edges represent various types of relations between pairs of entities. Given a knowledge graph composed of a collection of triplet facts $\Omega = \{(h, r, t)\}$, and a pre-defined embedding dimension $d$\footnote{To simplify the problem, we transform entities and relations into uniformly sized embedding spaces, \textit{i.e.} $d = k$.}, knowledge graph embedding aims to represent each entity $h \in E$ and relation $r \in R$ in a $d$-dimensional continuous vector space, where $E$ and $R$ are the sets of entities and relations, respectively. In other words, the embedding process projects textual triplets $(h, r, t)$ into a dense numerical vector space, where each entity or relation is transformed to a $d$-dimensional vector. With this vector representation, we can facilitate link prediction, DDI classification and other downstream applications. In drug-drug interaction knowledge graphs, drugs are entities and interactions between drugs are represented as relations. It is worth to note that all DDI knowledge graphs we introduced and utilized only contain the names of various drugs and interactions, and their direct relationships. Apart from that, there are no drug properties or other additional information available.

\begin{figure*}
    \centering
    \includegraphics[width=13cm]{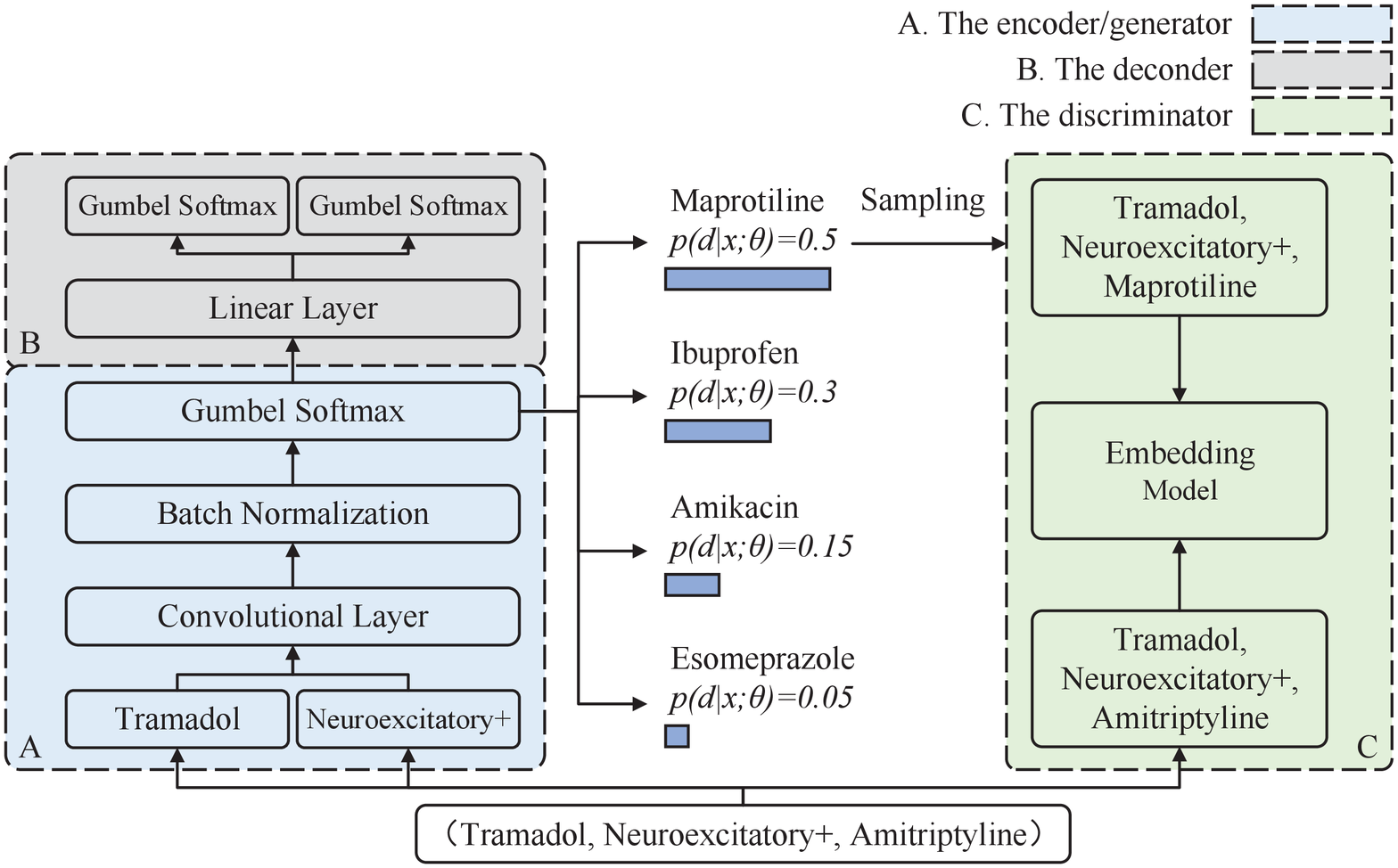}
    \caption{The architecture of the proposed adversarial autoencoder for knowledge graph embedding. (a) The encoder of the autoencoder learns to generate plausible negative triplets for the discriminator. (b) The decoder of the autoencoder is applied to further refine the performance of the encoder by minimizing reconstruction errors. (c) The generated negative triplet and the original positive triplet are both fed into the discriminator, as illustrated in the right part. (d) The discriminator is trained to yield a robust and effective knowledge graph representation model. Neuroexcitatory+ indicates the interaction ``increase neuroexcitatory activities''.}
    \label{Figure2}
\end{figure*}

Figure~\ref{Figure2} illustrates the proposed adversarial learning framework. At the beginning, a head or tail drug is discarded randomly from an authentic drug-drug interaction, and the resulting fragmentary triplet (Tramadol, increase neuroexcitatory activities, ?) is picked up as the input of the encoder. The encoder receives it and generates a one-hot vector which indicates another drug that has a similar effect or structure to Amitriptyline (such as Maprotiline which obtains the highest probability) from the collection of candidate drugs, this one-hot vector needs to be fed to both the decoder and the discriminator. For the decoder direction, the final outputs are two new vectors corresponding to the inputs of the encoder. The decoder restricts them to be as close to the inputs of the encoder as possible. As a consequence, we can not only guarantee that the model can generate different drugs, but also ensure that the generated drugs are proximal to the original ones in feature space. For the discriminator direction, the drug ``Maprotiline'' is selected to construct the final negative triplet (Tramadol, increase  neuroexcitatory activities, Maprotiline). Finally, the negative and positive triplets are jointly fed into the knowledge graph embedding model for learning embedding vectors.

\subsection{Autoencoder for sampling negative triplet facts}
The goal of the autoencoder is to provide more plausible negative triplets for the discriminator than what can be obtained via traditional random negative sampling.

\subsubsection{Shortcomings of traditional negative sampling}
Since Bordes et al.~\cite{transE2013} proposed to obtain corrupted triplets via uniform negative sampling, many researchers have applied this strategy to sample negative triples in the training process. This sampling strategy randomly selects a candidate entity from the entity set $E$ to replace the head or tail entity from the original positive triplet. It is worth to note that all candidate entities in entity set $E$ share the same probability of being drawn.

Obviously, this sampling method cannot contribute much to training an effective embedding model in most cases. As an example, given a valid triplet (\emph{Tramadol}, \emph{increase neuroexcitatory activities}, \emph{Amitriptyline}), our goal is to replace the tail drug with another acceptable drug to reassociate a plausible triplet. Given the word ``\emph{neuroexcitatory}'' in the relation and the drug type of ``\emph{Amitriptyline}'', it is intuitive to the domain expert that the tail drug should be a kind of pain reliever. If we choose the candidate drug in a random manner, many constructed negative triplets such as (\emph{Tramadol, increase neuroexcitatory activities, Esomeprazole}) or (\emph{Tramadol, increase the neuroexcitatory activities, Minoxidil}) can be trivially picked up as false by the discriminator, resulting in only infrequent parameter updates. By comparison, another generated triplet such as (\emph{Tramadol, increase neuroexcitatory activities, Acetaminophen}) seems to be a more reasonable DDI, because ``\emph{Acetaminophen}'' is more pharmacologically similar to ``\emph{Tramadol}'' than ``\emph{Minoxidil}''. It is worth noting that, although this traditional sampling method does not perform as well as the proposed method in most cases, when a knowledge graph is particularly sparse (in other words, there are extremely few triplets corresponding to each entity in the knowledge graph) the random sampling method has the opportunity to select any entity as a negative sampling option to train the model, so as to obtain a well-trained embedding model. This article does not consider such extreme cases.

As a consequence, we introduce an autoencoder to construct more plausible negative triplets instead of traditional uniform sampling. Here, the encoder aims to generate drugs as the generator in an adversarial learning framework, while the decoder restricts the manner and type of the generated drug, forcing it closer to the input drug and interaction. However, there is still a ``non-differentiability'' problem in discrete data generation.

\subsubsection{Gumbel-Softmax relaxation with discrete data}
In this section, we first illustrate why training adversarial learning models with discrete data is a vital issue. From a mathematical perspective, assuming the total number of drugs (entities) is $\left | E \right |$, the next generated one-hot index vector $y \in \mathbb{R}^{\left | E \right |}$ can be obtained by sampling:
\begin{equation}
\label{equ12}
y \sim \sigma\left(o\right),
\end{equation}
where $o \in \mathbb{R}^{\left | E \right |}$ denotes the output logits of the last layer in the generator, $\sigma(\cdot)$ indicates the Softmax function. The sampling operation in Equation~\eqref{equ12} implies a step function that is not differentiable at the end of the generator output. Because the differential coefficient of a step function is $0$ almost everywhere, we have $\frac{\partial y}{\partial \theta_{G}} = 0, \ a.e.$, where $\theta_{G}$ are the parameters of the generator. According to the chain rule, the gradients of the generator loss $l_{G}$ with respect to $\theta_{G}$ are calculated as:
\begin{equation}
\frac{\partial l_{G}}{\partial \theta_{G}}= \frac{\partial y}{\partial \theta_{G}} \frac{\partial l_{G}}{\partial y}=0 \quad \ a.e.
\end{equation}

As a result, $\partial l_{G}/ \partial \theta_{G} = 0$ means that the gradients of the generator loss cannot be propagated back to the generator via the discriminator. In other words, the generator cannot update its own parameters based on the feedback provided by the discriminator. This phenomenon is called the ``vanishing gradient'' or ``non-differentiability'' issue of adversarial learning models in discrete data domains. 

From an instance point of view, even though a Softmax output vector of the generator $\alpha = [0.25, 0.35, 0.25, 0.15]$ can improve the performance of the generator to optimize $\alpha$ to $\beta = [0.05, 0.70, 0.15, 0.10]$ allowing localizing a specific entity, the final sampling result has not changed, \textit{i.e.}, $Onehot(\alpha) = Onehot(\beta) = [0,1,0,0]$. The identical sampling one-hot vectors are repeatedly fed to the discriminator, so that the gradients obtained by the discriminator are inoperative, and the convergence direction of the generator is indistinct, no matter how powerful the discriminator may be.

In order to solve the ``non-differentiablity'' issue, this paper leverages a Gumbel-Softmax relaxation technique which can approximate patterns sampled from a categorical distribution by defining a continuous distribution on the simplex. There are two parts in the Gumbel-Softmax relaxation: $(1)$ The Gumbel-Max trick. Following previous studies proposed by Jang et al.~\cite{jang2016categorical} and Maddison et al.~\cite{maddison2016concrete}, the sampling in Equation~\eqref{equ12} should be reparametrized as:
\begin{equation}
\label{equ15}
y=\text {one\_hot }\left(\mathop{\arg\max}_{1 \leq i \leq \left | E \right |}\left(o_{i}+g_{i}\right)\right),
\end{equation}
where $o_{i}$ is the $i$-th element of $o$ and $g_{1}, \cdots, g_{\left | E \right |}$ are i.i.d.\ samples drawn from a standard Gumbel distribution, \textit{i.e.}, $g_{i} = \log(-\log U_{i})$ with $U_{i} \sim Uniform (0,1)$. $(2)$ Relaxing the discreteness. So far the ``$\arg\max$'' operation in Equation~\eqref{equ15} is still non-differentiable. We employ the Softmax function as a differentiable, continuous approximation to further approximate it, and calculate a $\left | E \right |$-dimensional sample vector $\hat{y}$. Each entry in $\hat{y}$ is acquired by:
\begin{equation}
\hat{y}_{i}=\frac{\exp \left(\left( \left(o_{i}\right)+g_{i}\right) / \tau\right)}{\sum_{a=1}^{\left | E \right |} \exp \left(\left(\left(o_{a}\right)+g_{a}\right) / \tau\right)}, %\text{for} \quad a=1, \dots, D
\end{equation}
where $\tau > 0$ is an adjustable parameter referred to as the \textit{inverse temperature}. When the temperature $\tau$ approaches $0$, the sampled vectors from the Gumbel-Softmax distribution are equal to one-hot vectors and the Gumbel-Softmax distribution becomes identical to the categorical distribution. It is worth noting that, in this way, $\hat{y}$ can be differentiated with respect to $o$, we can utilize $\hat{y}$ to replace one-hot vector $y$ as the final output of the generator. 
Consequently, the ``non-differentiability'' issue is solved by taking advantage of the Gumbel-Softmax relaxation. The generator (the encoder part of our autoencoder) can smoothly generate one-hot vectors that indicate plausible drugs.
%and this generator combines with the decoder is autoencoder, combines with the discriminator is generative adversarial network. 
\subsubsection{Autoencoder architecture}
In the generator, each drug and interaction are initially transformed from a one-hot index vector to a specific embedding feature space associated with two embedding matrices, one for drugs, indicated by $E^{\left | E \right | \times d}$, and one for interactions, indicated by $R^{\left | R \right | \times k}$, $\left | E \right |$ and $\left | R \right |$ are the total numbers of drugs and interactions, respectively. In this paper, the embedding dimensionality of drugs is identical to that of interactions, \textit{i.e.} $d = k$. Because of this setting, we can concatenate the embedded vectors of head drug $\bm{h}$ and interaction $\bm{r}$, and reshape it as an input $\bm{A} = Reshape([\bm{h},\bm{r}])$ to the 2D convolutional network layer which has been shown to extract available features with filters $\omega$. A feature map tensor $\mathcal{T} \in \mathbb{R}^{b \times m \times n}$ is calculated through this layer, where $b$ is the number of feature maps with dimensions $m$ and $n$. After that, we reshape the tensor $\mathcal{T}$ into a single vector $\bm{t} \in \mathbb{R}^{bmn}$, and then transform it into an $\left | E \right |$-dimensional feature vector by using the projection matrix $\mathbf{W} = \mathbb{R}^{bmn \times \left | E \right |}$. Finally, the Gumble-Softmax relaxation described above is applied to generate a plausible tail drug. Mathematically, the one-hot vector of drug $\bm{v}$ is calculated as:
\begin{equation}
    \bm{v}= g \left(Re(Re([\bm{h};\bm{r}] *\omega)\mathbf{W}) \right),
\end{equation}
where $Re(\cdot)$ represents the reshape operation, and $g(\cdot)$ is the Gumble-Softmax relaxation. The output of the generator is a one-hot index vector that refers to a specific drug. This drug, when associated with the inputs of the generator including head drug and interaction forms the corrupted triplet.

The one-hot vector $\bm{v}$ acts as the input, given by two linear network layers. In order to invoke the restriction of the autoencoder forcing the neural network to capture only significant features of the data, there are two outputs in the decoder. These two output dimensions are $\left | E \right |$ and $\left | R \right |$, corresponding to the dimensions of the two generator inputs.

\subsection{Knowledge graph embedding discriminator}
The discriminator used in our framework is constructed following previous research. As described in Section~\ref{section2}, the individual models have different structures and score functions. The embeddings of drugs and interactions are obtained by minimizing the ranking loss associated with positive and negative triplets. Different from previous models in which negative samples are generated via random sampling from whole drug set, we apply an autoencoder to construct more plausible triplets to refine the performance of the model.

\subsection{Training strategy}
The training procedure is comprised of three main parts: \textbf{i}) the parameter update of the autoencoder in which $G$ and $A$ indicate the generator (the encoder) and the decoder, respectively; \textbf{ii}) the parameter update of the discriminator $D$; \textbf{iii}) the parameter update of the generator $G$.

The autoencoder is designed to learn an effective representation of data. In this paper, we employ the encoder network $G(\bm{z};\theta)$ to generate high-quality negative drugs and apply the decoder network $A(\bm{x}; \eta)$ to restrict the sampling direction to obtain more plausible samples. To update parameters $\theta$ and $\eta$, we train the autoencoder by minimizing the reconstruction error $L_{G,A}$:
\begin{equation}
\min L_{G,A} = \min _{(\mu = \theta, \eta)} \left \|  x - A\left(G\left(z; \theta\right) ; \eta\right) \right \|^{2}.
\end{equation}

The goal of the discriminator network $D(\bm{x}; \phi)$ is to distinguish a sample $\bm{x}$ as originating either from the real distribution $p_{r}(\bm{x})$ or the generator $p_{\theta}(\bm{x})$. Given an original training sample $(\bm{x}, y)$, $y \in \{1, -1\}$ signals whether it is a true sample from $p_{r}(\bm{x})$ or a generated sample from $p_{\theta}(\bm{x})$, the optimization objective of the discriminator $L_{D}$ is to minimize cross-entropy:
\begin{equation}
\label{equ18}
\min L_{D} = \min _{\phi}-\left(y \log p(y=1 | \bm{x})+(1-y) \log p(y=0 | \bm{x})\right).
\end{equation}

If the distribution $p(\bm{x})$ is a mixture of distributions $p_{r}(\bm{x})$ and $p_{\theta}(\bm{x})$ in equal proportions, \textit{i.e.}, $p(\bm{x}) = \frac{1}{2}(p_{r}(\bm{x}) + p_{\theta}(\bm{x}))$, then Equation~\eqref{equ18} can be rewritten as:
\begin{equation}
\min L_{D} = \min _{\phi} -\left (\log D(\bm{x}; \phi)+\log (1-D(G(z^{(i)} ; \theta) ; \phi)) \right ).
\end{equation}

The goal of the generator is the opposite of the discriminator, the generator tries to construct negative samples which can fool the discriminator into confusing a negative sample for a real one. Its objective function $L_{G}$ is formulated as:
\begin{equation}
\min L_{G} = \min _{\theta}\left(\log (1-D(G(\bm{z}; \theta), \phi))\right).
\end{equation}

Compared with a traditional single-objective optimization task, the optimization goals of these two networks in the adversarial game are extremely challenging. There are many potential issues in the traditional adversarial network training process such as training instability and difficulty, uninformative loss functions of generator and discriminator, and a lack of diversity in the generated samples.

These problems are caused by the attempt to minimize the Jensen–Shannon (JS) divergence  between the real distribution $p_{r}(\bm{x})$ and the generated distribution $p_{\theta}(\bm{x})$. The JS divergence can only be computed when two distributions $P$, $Q$ have overlapping parts. When these two distributions do not overlap or the overlapping parts are negligible in size, their JS divergence is identically equal to $\log 2$. That means that when the real distribution $p_{r}(\bm{x})$ and the generated distribution $p_{\theta}(\bm{x})$ have no overlap, the outputs of the discriminator are $0$ for all generated data, \textit{i.e.} $D(G(\bm{z}, \theta)) = 0, \forall \bm{z}$. As a result, the gradients of the generator vanish.

Inspired by Wasserstein GANs~\cite{arjovsky2017wasserstein} in which Wasserstein distance (also known as Earth-Mover distance) is introduced as a more robust metric to replace the JS divergence, we use this distance measure to improve the performance of our knowledge graph embedding framework in this article. Given a real distribution $p_{r}(\bm{x})$ and a generated distribution $p_{\theta}(\bm{x})$, the 1st-Wasserstein distance between them is formalized as:
\begin{equation}
\label{equ22}
    W\left(p_{r}, p_{\theta}\right)=\inf _{\gamma \sim \Pi\left(P_{r}, P_{\theta}\right)} \mathbb{E}_{(\mathbf{x}, \mathbf{y}) \sim \gamma}[\|\mathbf{x}-\mathbf{y}\|],
\end{equation}
where $\Pi\left(P_{r}, P_{\theta}\right)$ is the set of all possible joint distributions with marginal distribution $\gamma(\bm{x},\bm{y})$. When there are no overlapping or slightly overlapping parts between two distributions, the JS divergence becomes constant. In contrast, the 1st-Wasserstein distance can measure distances between two non-overlapping distributions.

Equation~\eqref{equ22} is difficult to calculate directly, and needs to be transformed into a solvable form via the Kantorovich-Rubinstein duality theorem~\cite{villani2008optimal}. 
According to this theorem, the Wasserstein distance between two distributions can be converted into an upper bound on the expected difference between distributions $p_{r}$ and $p_{\theta}$ for a function that satisfies the K-Lipschitz continuum. We rewrite the 1st-Wasserstein distance:
\begin{equation}
\label{equ23}
    \bm{W}\left(p_{r}, p_{\theta}\right)=\frac{1}{K} \sup _{\|f\|_{L} \leq K}\left(\mathbb{E}_{x \sim p_{r}}[f(\bm{x})]-\mathbb{E}_{x \sim p_{\theta}}[f(\bm{x})]\right),
\end{equation}
where $f(\cdot)$ is the K-Lipschitz function, that satisfies:
\begin{equation}
    \|f\|_{L}  \sup _{x \neq y} \frac{|f(\bm{x})-f(\bm{y})|}{|\bm{x}-\bm{y}|} \leq K.
\end{equation}

If a function is differentiable and its derivatives are bounded, then this function is a Lipschitz continuous function. Because the discriminator neural network $D(x;\phi)$ satisfies the above conditions, it is also a Lipschitz continuous function, allowing us to approximate the upper bound in Equation~\eqref{equ23} as:
\begin{equation}
    \min _{\phi}-\left(\mathbb{E}_{x \sim p_{r}}[D(\bm{x} ; \phi)]-\mathbb{E}_{x \sim p_{\theta}}[D(\bm{x} ; \phi)]\right).
\end{equation}

Different from standard discriminator networks in which the final layer is a sigmoid function over an output range of $[0, 1]$, at this point, we only need to find a network $D({\bm{x}; \phi})$ that maximizes the difference in expectations between the two distributions $p_{r}$ and $p_{\theta}$. As a consequence, the final layer in our discriminator $D({\bm{x}; \phi})$ is a linear layer, and its range is not limited. That means that, for real samples, the score of $D({\bm{x}; \phi})$ should be high, and for samples generated by the model, low scores are expected.

Moreover, to make $D({\bm{x}; \phi})$ satisfy the K-Lipschitz continuity condition, we can approximate it by limiting the range of the parameter $\phi$, such that $\phi \in [-c,c]$, $c$ is a relatively small positive number.

The goal of the generator is to minimize the Wasserstein distance, making the real distribution $p_{r}$ and the generated distribution $p_{\theta}$ coincide as much as possible, \textit{i.e.}:
\begin{equation}
    \min _{\theta} - \mathbb{E}_{z \sim p(z)}[D(G(z ; \theta) ; \phi)].
\end{equation}

Because $D(\bm{x}; \phi)$ is an unsaturated function, the gradients of the generator parameters $\theta$ will not disappear. This solves the problem of instability in original adversarial framework. In addition, by replacing the JS divergence by the Wasserstein distance, the objective function of the generator in this framework can alleviate the model collapse problem to a certain extent and make the generated samples more diverse. The detailed procedure of this adversarial framework for knowledge graph embedding is described in Algorithm~\ref{al1}.
\begin{algorithm}
\caption{Adversarial training of the autoencoder with Wasserstein distance for learning interactions in knowledge graph embeddings.}
\label{al1}
\SetKwInOut{Input}{Input}\SetKwInOut{Output}{Output}

\Input{The set of positive DDIs $T = \{(h, r, t)\}$, the number of training iterations $e$, the number of discriminator iterations per generator iteration $n_{dis}$, mini-batch size $m$, the learning rate of the generator $\alpha$, the learning rate of the discriminator $\beta$, the clipping parameter $c$.}
\Output{Drugs and interactions embeddings learned by $D$.}
\BlankLine
\textbf{Initialize} the generator $G$ with parameters $\theta_{0}$, the decoder $A$ with parameters $\eta_{0}$, the discriminator $D$ with parameters $\phi_{0}$\;
\For{$k = 1, \cdots, e$}{
\For{$i = 1, \cdots, m$}{
$L^{(i)}_{G,A}=  \left \| x^{(i)} - A\left(G\left(z^{(i)} ; \theta\right) ; \eta\right) \right \|^{2}$ \tcp*[l]{Update the autoencoder by minimizing $L^{(i)}_{G,A}$}
} 
$g_{\mu} \leftarrow \nabla_{\mu} \frac{1}{m} \sum_{i=1}^{m} L^{(i)}_{G,A}, \quad (\mu = \theta, \eta)$\;
$\theta \leftarrow \theta - \alpha \cdot \operatorname{Adagrad}\left(\theta, g_{\mu}\right)$\;
$\eta \leftarrow \eta - \alpha \cdot \operatorname{Adagrad}\left(\eta, g_{\mu}\right)$\;
\For{$t = 1, \cdots, n_{dis}$}{
    \For{$i = 1, \cdots, m$}{
    $L^{(i)}_{D} = - \left[D\left(x^{(i)} ; \phi\right)-D\left(G\left(z^{(i)} ; \theta\right) ; \phi\right) \right]$ \tcp*[l]{Update the discriminator by minimizing $L^{(i)}_{D}$}
    }
	$f_{\phi} \leftarrow \nabla_{\phi} \frac{1}{m} \sum_{i=1}^{m} L^{(i)}_{D}$\;
	$\phi \leftarrow \phi - \beta \cdot \operatorname{Adagrad}\left( \phi, f_{\phi} \right)$\;
	$\phi \leftarrow \operatorname{clip}(\phi,-c, c)$\;}
\For{$i = 1, \cdots, m$}{
    $L^{(i)}_{G} = -D\left(G\left(z^{(i)} ; \theta\right) ; \phi\right)$\tcp*[l]{Update the generator by minimizing $L^{(i)}_{G}$}
}
$h_{\theta} \leftarrow \nabla_{\theta} \frac{1}{m} \sum_{i=1}^{m} L^{(i)}_{G}$\;
$\theta \leftarrow \theta - \alpha \cdot \operatorname{Adagrad}\left( \theta, h_{\theta} \right)$\;
}
\textbf{Return} Drug and interaction embeddings.
\end{algorithm}

%%%%%%%%%%%%%%%%%%%%%%%%%%%%%%%%%%%%%%%%%%%%%%%%%%%%%%%%%%%%%%%%%%%%%%%%%%%%%%%%%%%%%
\section{Experiments}
\label{section4}
In this section, we first describe the experimental datasets, then introduce important hyper-parameter settings and comparison methods for our experiment. Afterwards, link prediction and DDI classification experiments are constructed for comparing performance of the proposed methods with benchmark and the state-of-the-art models. Finally, we project the high-dimensional embedding feature space to two-dimensions for visual inspection of qualitative example outputs.

\subsection{Datasets}
We conduct our link prediction and DDI classification experiments on two widely used public datasets: \textbf{DeepDDI} and \textbf{Decagon}. For both datasets, we randomly sample 80\% of drug-drug pairs as training data, 10\% as validation data and the remaining 10\% as test data. Statistics of these two datasets are collected in Table~\ref{table1}.

\begin{table}[htbp]
\caption{Statistics of the data sources}
\centering
\setlength{\tabcolsep}{1.8mm}{
\begin{tabular}{l|ccccc}
\hline
Datasets & \#Drugs & \#DDI Types &\#Train & \# Valid & \#Test \\ \hline
DeepDDI & 1,710 & 86 & 153,828 & 19,228 & 19,228 \\
Decagon & 637 & 200 & 897,446 & 112,181 & 112,181 \\ \hline
\end{tabular}}
\label{table1}
\end{table}

\textbf{DeepDDI}~\cite{ryu2018deep} is composed of 1,710 drugs and 86 different interaction types from DrugBank~\cite{wishart2008drugbank} capturing 192,284 drug-drug pairs as samples. 99.87\% of drug-drug pairs only have one type of DDI.

\textbf{Decagon}~\cite{zitnik2018modeling} is composed of 637 drugs and 200 different interaction types from the TWOSIDES dataset~\cite{tatonetti2012data} capturing 1,121,808 drug-drug pairs as samples. We follow common practice by sampling 200 medium frequency DDI types ranging from Top-600 to Top-800, ensuring that every DDI type has at least 90 drug combinations. 73.27\% of drug-drug pairs have more than one type of DDI.

\subsection{Comparison methods}
To comprehensively evaluate the performance of our proposed model, we select several representative methods from the three categories of knowledge graph embedding as baselines to be compared with our approach. These baselines are described as follows:
\begin{itemize}
    \item \textbf{TransE}~\cite{transE2013} represents both entities and relations in a low-dimensional feature space, and interprets relations as translation operations to concatenate the entities.
    \item \textbf{DistMult}~\cite{distmult2015} proposes a multi-relational learning method in which the bilinear objective is effective at capturing relational semantics.
    \item \textbf{ComplEx}~\cite{complex2016} describes a simple tensor factorization method using embedding vectors with complex values to handle symmetric and asymmetric relations.
    \item \textbf{KBGAN}~\cite{kbgan2018} introduces an adversarial learning framework for knowledge graph embedding in which a generator is applied to sample negative triplets for refining the performance of the discriminator. 
    \item \textbf{SimplE}~\cite{kazemi2018simple} develops an embedding method based on Canonical Polyadic decomposition that extends the model to learn the two embedding vectors of each entity independently.
    \item \textbf{RotatE}~\cite{sun2018rotate} embeds entities as complex-value vectors and defines relations as rotations from the head entity to the tail in a complex vector space. In addition, it utilizes a new self-adversarial negative sampling method to train the embedding model.
\end{itemize}

\subsection{Link prediction}
Link prediction is a characteristic task which aims to infer the missing drug when given an existing drug and interaction query. Concretely, the target of link prediction is to predict the missing drug $t$ if given $(h, r)$ or predict $h$ given $(r, t)$. Results are obtained via ranking by discriminator scores.

\subsubsection{Metrics}
For each drug-drug interaction $(h, r, t)$ in the test set, the real head drug (or tail drug) is circularly replaced by all drugs in the drug set $E$. Then, the scores corresponding to all triplets are computed, all scores are ranked in descending order.

However, some reconstructed DDIs might coincidentally be authentic in the DDI knowledge graph. In this case, the reconstructed DDI which is a true fact might yield a high ranking, resulting in an inaccurate assessment. To avoid this situation, following Bordes et al.~\cite{transE2013}, we employ the ``\textbf{Filtered}'' setting to eliminate all reconstructed DDIs which appear either in the training, validation, or test datasets. Finally, model performance is measured in terms of:
\begin{itemize}
    \item \textbf{MR}: the average rank of the real entities.
    \item \textbf{MRR}: the average reciprocal rank of the real entities.
    \item \textbf{HITS@N\%}: the proportion of real entities that ranked in the top $N$. Here, we specially choose $N = 1, 3, 10$ to validate the performance of compared models.
\end{itemize}

It should be noted that good performance is indicated by low \textbf{MR} and high \textbf{MRR} and \textbf{HITS@N\%} scores.

\subsubsection{Training protocol}
We utilize the Adagrad self-adaptive optimizer for training, and perform parameter optimization via limited grid search: the learning rate of the generator $\alpha\in\{0.01, 0.005, 0.001\}$, the learning rate of the discriminator $\beta\in\{0.5, 0.1, 0.05, 0.01\}$, the size of drug and interaction embedding vectors $d \in\{50, 100, 200\}$, the mini-batch size $m\in\{256, 512, 1024\}$, the number of discriminator training iterations per generator iteration $n_{dis}\in\{1, 2, 5\}$ and the number of overall training iterations $e\in\{300, 500, 700, 1000\}$. The final parameter settings are determined on the validation set. 

On the DeepDDI dataset, the best configurations are \{$\alpha$ = 0.001, $\beta$ = 0.05, $d$ = 200, $m$ = 512, $n_{dis}$ = 1, $e$ = 300\} for our model with ComplEx, \{$\alpha$ = 0.001, $\beta$ = 0.1, $d$ = 200, $m$ = 512, $n_{dis}$ = 1, $e$ = 300\} for our model with SimplE and \{$\alpha$ = 0.001, $\beta$ = 0.5, $d$ = 200, $m$ = 512, $n_{dis}$ = 2, $e$ = 500\} for our model with RotatE. On the Decagon dataset, the best configurations are \{$\alpha$ = 0.005, $\beta$ = 0.5, $d$ = 200, $m$ = 1024, $n_{dis}$ = 1, $e$ = 1000\} for our model with ComplEx, \{$\alpha$ = 0.005, $\beta$ = 0.5, $d$ = 200, $m$ = 512, $n_{dis}$ = 2, $e$ = 1000\} for our model with SimplE and \{$\alpha$ = 0.005, $\beta$ = 0.5, $d$ = 200, $m$ = 512, $n_{dis}$ = 5, $e$ = 1000\} for our model with RotatE..

\begin{table*}[]
\renewcommand\arraystretch{1.2}
\caption{Evaluation results on link prediction. The best two results are highlighted in \textcolor[rgb]{1,0,0}{red}, \textcolor[rgb]{0,0,1}{blue} and are formatted in boldface.}
\resizebox{\textwidth}{!}{
\begin{tabular}{l|ccccc|ccccc}
\hline
\multirow{2}{*}{\textbf{Methods}} & \multicolumn{5}{c|}{\textbf{DeepDDI}} & \multicolumn{5}{c}{\textbf{Decagon}} \\ \cline{2-11} 
 & MR & MRR & HITS@1\% & HITS@3\% & HITS@10\% & MR & MRR & HITS@\%1 & HITS@3\% & HITS@10\% \\ \hline
TransE\cite{transE2013} & 33.1338 & 0.2355 & 0.0069 & 0.3858 & 0.6216 & 44.7126 & 0.1005 & 0.0001 & 0.1106 & 0.2923 \\
DistMult\cite{distmult2015} & 42.9008 & 0.1474 & 0.0532 & 0.2063 & 0.3765 & 57.9994 & 0.1212 & 0.0415 & 0.1149 & 0.2506 \\
ComplEx\cite{complex2016} & 15.8439 & 0.5196 & 0.3731 &0.5725 & 0.7889 & 20.6787& 0.2133 & 0.0514 & 0.2018 & 0.4249 \\
KBGAN\cite{kbgan2018} & 15.6894 & 0.5209 & 0.3824 & 0.5791 & 0.7957 & \textcolor[rgb]{0,0,1}{\textbf{20.2376}} & 0.2068 & 0.0492 & 0.1893 & 0.4081 \\
SimplE\cite{kazemi2018simple} & 16.0407 & 0.5037 & 0.3650 & 0.5738 & 0.7895 & 21.3255 & 0.2034 & 0.0484 & 0.1906 & 0.4188 \\
RotatE\cite{sun2018rotate} & 15.5169 & 0.4925 & 0.3432 & 0.5624 & 0.7683 & 43.0815 & 0.1705 & \textcolor[rgb]{0,0,1}{\textbf{0.0928}} & 0.1686 & 0.3203\\
Our model (ComplEx) & \textcolor[rgb]{1,0,0}{\textbf{11.9863
}} & \textcolor[rgb]{1,0,0}{\textbf{0.5455}} & \textcolor[rgb]{1,0,0}{\textbf{0.4076}} & \textcolor[rgb]{1,0,0}{\textbf{0.6224}} & \textcolor[rgb]{0,0,1}{\textbf{0.8046}} & \textcolor[rgb]{1,0,0}{\textbf{19.9381}} & \textcolor[rgb]{1,0,0}{\textbf{0.2280}} & 0.0613 & \textcolor[rgb]{1,0,0}{\textbf{0.2201}} & \textcolor[rgb]{1,0,0}{\textbf{0.4372}} \\
Our model (SimplE) & 14.5913 & \textcolor[rgb]{0,0,1}{\textbf{0.5308}} & \textcolor[rgb]{0,0,1}{\textbf{0.3979}} & 0.6019 & 0.7996 & 20.6682 & \textcolor[rgb]{0,0,1}{\textbf{0.2176}} & 0.0539 & \textcolor[rgb]{0,0,1}{\textbf{0.2047}} & \textcolor[rgb]{0,0,1}{\textbf{0.4287}}\\
Our model (RotatE) & \textcolor[rgb]{0,0,1}{\textbf{13.8462}} & 0.5284 & 0.3904 & \textcolor[rgb]{0,0,1}{\textbf{0.6044}} & \textcolor[rgb]{1,0,0}{\textbf{0.8071}} & 41.3432 & 0.1772 & \textcolor[rgb]{1,0,0}{\textbf{0.0974}} & 0.1763 & 0.3314 \\\hline
\end{tabular}}
\label{table2}
\end{table*}

\subsubsection{Comparison with state-of-the-art models}
We adopt the above configurations to train our model and  compare our results to state-of-the-art methods. Table~\ref{table2} shows a detailed comparison of the proposed approach and comparative methods on the two standard benchmark datasets. We can observe that:
\begin{itemize}
    \item On both datasets, the knowledge graph embedding models trained via our proposed adversarial framework obtain a better performance on all metrics compared with other state-of-the-art methods. Especially ComplEx model we trained, it achieved the best performance on both datasets.
    
    \item As early models in KG embedding, TransE and DistMult have their inherent limitations in expressiveness compared with current methods. These issues are unlikely to be completely compensated by advanced training approaches. That is the reason against training them via adversarial learning in the experiment. And ComplEx and SimplE, which can be regarded as extension version of DistMult, achieve an improvement on these metrics by introducing complex-valued embeddings.
    
    \item Compared with these models with uniform negative sampling, KBGAN and RotatE also yield good performances on the two benchmark datasets. Especially, KBGAN achieves the second best result in MR and RotatE achieves the second best result in HITS@1\% on the Decagon dataset. The pivotal factor to their good performance is that they also utilize adversarial learning to train the model. More detailed analyses are provided in Section~\ref{section5.1}.
    
    \item On the DeepDDI datasets, the proposed framework can improve the performance by an average of 3 points of HITS@10\% beyond the original methods. Even under the increased complexity of the Decagon dataset that includes more interaction types and 73.27\% drug-drug pairs having more than one type of interaction, we observe an average improvement of 1 percentage point.
\end{itemize}

\subsection{DDI classification}
DDI classification is an important pharmacological task that aims to determine the authenticity of a DDI triplet. As some existing articles~\cite{ryu2018deep,ma2019genn,karim2019drug} investigate DDI prediction, we follow them in casting DDI classification as a multi-label interaction prediction problem. 

Given a pair of drugs, we first construct DDI triplets by repeatedly adding each interaction in the interaction set $R$ into the pair of drugs, then estimate the confidence in every generated triplet. Those interactions corresponding to high-score triplets are the ones we want to obtain.
\subsubsection{Metrics}
\begin{itemize}
    \item \textbf{ROC-AUC}: the area under the receiver operating characteristic curve.
    \item \textbf{PR-AUC}: the area under the precision-recall curve.
    \item \textbf{P@K}: the mean percentage of true predicted labels among TOP-K over all samples. In this paper, $K = 1, 3, 5$ are selected as evaluation indicators to estimate the performance of models.
\end{itemize}

\subsubsection{Training protocol}
\label{subsection4.4.2}
In this task, we use the models trained for link prediction. Thus, all settings and hyper-parameter configurations are retained from above.
\begin{table*}[h]
\renewcommand\arraystretch{1.2}
\caption{Evaluation results on DDI classification. The best two results are highlighted in \textcolor[rgb]{1,0,0}{red}, \textcolor[rgb]{0,0,1}{blue} and are formatted in boldface.}
\resizebox{\textwidth}{!}{
\begin{tabular}{l|ccccc|ccccc}
\hline
\multirow{2}{*}{\textbf{Methods}} & \multicolumn{5}{c|}{\textbf{DeepDDI}} & \multicolumn{5}{c}{\textbf{Decagon}} \\ \cline{2-11} 
 & PR-AUC & ROC-AUC & P@1 & P@3 & P@5 & PR-AUC & ROC-AUC & P@1 & P@3 & P@5 \\ \hline
ComplEx\cite{complex2016} & 0.7419 & 0.9355 & 0.6866 & 0.2537
 & 0.1597 & 0.3568 & 0.9391 & 0.2212 & 0.1852 & 0.1618 \\
KBGAN\cite{kbgan2018} & 0.7562 & 0.9436 & 0.6919 & 0.2643 & 0.1612 & 0.3605 & 0.9414 & 0.2270 & 0.1869 & 0.1637 \\
SimplE\cite{kazemi2018simple} & 0.7499 & 0.9310 & 0.6927 & 0.2553 & 0.1625 & 0.3588 & 0.9396 & 0.2241 & 0.1857 & 0.1623 \\
RotatE\cite{sun2018rotate} & 0.7676 & 0.9348 & 0.7084 & 0.2678 & \textcolor[rgb]{0,0,1}{\textbf{0.1668}} & 0.2138 & 0.8390 & 0.1449 & 0.1168 & 0.1020 \\
Our model (ComplEx) & 0.7615 & \textcolor[rgb]{1,0,0}{\textbf{0.9527}} & 0.7030 & 0.2631 & 0.1644 & \textcolor[rgb]{0,0,1}{\textbf{0.3623}} & \textcolor[rgb]{0,0,1}{\textbf{0.9469}} & \textcolor[rgb]{1,0,0}{\textbf{0.2306}} & \textcolor[rgb]{0,0,1}{\textbf{0.1934}} & \textcolor[rgb]{0,0,1}{\textbf{0.1685}} \\
Our model (SimplE) & \textcolor[rgb]{0,0,1}{\textbf{0.7693}} & 0.9431 & \textcolor[rgb]{0,0,1}{\textbf{0.7177}} & \textcolor[rgb]{0,0,1}{\textbf{0.2685}} & 0.1661 & \textcolor[rgb]{1,0,0}{\textbf{0.3642}} & \textcolor[rgb]{1,0,0}{\textbf{0.9480}} & \textcolor[rgb]{0,0,1}{\textbf{0.2291}} & \textcolor[rgb]{1,0,0}{\textbf{0.1935}} & \textcolor[rgb]{1,0,0}{\textbf{0.1703}} \\
Our model (RotatE) & \textcolor[rgb]{1,0,0}{\textbf{0.7899}} & \textcolor[rgb]{0,0,1}{\textbf{0.9480}} &\textcolor[rgb]{1,0,0}{\textbf{0.7296}} & \textcolor[rgb]{1,0,0}{\textbf{0.2762}} & \textcolor[rgb]{1,0,0}{\textbf{0.1722}} & 0.2215 & 0.8485 & 0.1501 & 0.1216 & 0.1059 \\\hline
\end{tabular}}
\label{table3}
\end{table*}
\subsubsection{Comparison with state-of-the-art models}
The drug-drug interaction classification results are displayed in Table~\ref{table3}. Since TransE and DistMult are comparably primitive models, their performance is not expected to be convincing for this task and the corresponding models are not included in this experiment. As can be seen from Table~\ref{table3}:
\begin{itemize}
    \item Similar to the link prediction task, our proposed method achieves consistent improvements in this scenario. On both datasets, the proposed framework refines the performance of all baseline knowledge graph embedding models. 
    
    \item Where the improved ComplEx method outperformed other models on link prediction, for the classification task, the improved RotatE method yields the best results on the DeepDDI dataset while the improved SimplE method obtains the best performance on Decagon. 
    
    \item The performance of RotatE is highly variable across datasets. The improved model yields the best performance in four of the five metrics on the DeepDDI dataset (PR-AUC, P@1, P@3 and P@5). Even the P@5 results rank a close second. However, on the Decagon dataset, RotatE and its adversarial training version show the worst results among all methods. We will discuss the specific reasons for these results in Section~\ref{section5.2}.
    
    \item On the DeepDDI datasets, the proposed framework can improve the performance by an average of 2 points of PR-AUC beyond the original methods. Even under the increased complexity of the Decagon dataset, we observe an average improvement of 0.6 percentage point.
\end{itemize}
%%%%%%%%%%%%%%%%%%%%%%%%%%%%%%%%%%%%%%%%%%%%%%%%%%%%%%%%%%%%%%%%%%%%%%%%%%%%%%%%%%%%%
\section{Discussion}
\label{section5}
\begin{table*}[]
\caption{Some instances of negative samples constructed by random and generator sampling from the DeepDDI dataset. In this table, all drugs are shown in bold and all interactions are denoted by star. Additionally, there is a parenthesis below each drug, which contains the function or character of the drug. The triplets in the first column are positive, the underlined drug signifies that it would be replaced by other drugs in the next two columns. The replacement drugs sampled randomly are listed in the second column, and the third column displays the drugs generated by our method.}
\resizebox{\textwidth}{!}{
\begin{tabular}{c|c|c}
\hline
\textbf{Positive triplets} & \textbf{Random sampling} & \textbf{Generator sampling} \\ \hline
\begin{tabular}[c]{@{}c@{}}\textbf{Midazolam}\\ (hypnotic sedative)\end{tabular} & \begin{tabular}[c]{@{}c@{}}\textbf{Lumefantrine}\\ (antimalarial)\end{tabular} & \begin{tabular}[c]{@{}c@{}}\textbf{Methadyl acetate}\\ (narcotic analgesic)\end{tabular} \\
 increases the risk of adverse effects & \begin{tabular}[c]{@{}c@{}}\textbf{Diltiazem}\\ (antihypertensive)\end{tabular} & \begin{tabular}[c]{@{}c@{}}\textbf{Levacetylmethadol}\\ (narcotic analgesic)\end{tabular} \\
\begin{tabular}[c]{@{}c@{}}\textbf{\underline{Dezocine}}\\ (partial opiate)\end{tabular} & \begin{tabular}[c]{@{}c@{}}\textbf{Pefloxacin}\\ (antibacterial)\end{tabular} & \begin{tabular}[c]{@{}c@{}}\textbf{Nalbuphine}\\ (narcotic)\end{tabular} \\ \hline
\begin{tabular}[c]{@{}c@{}}\textbf{Treprostinil}\\ (treatment of pulmonary hypertension)\end{tabular} & \begin{tabular}[c]{@{}c@{}}\textbf{Carmustine}\\ (treatment of brain tumors)\end{tabular} & \begin{tabular}[c]{@{}c@{}}\textbf{Ridogrel}\\ (prevention of thrombo-embolism)\end{tabular} \\
 increases the antiplatelet activities & \begin{tabular}[c]{@{}c@{}}\textbf{Plicamycin}\\ (antineoplastic antibiotic)\end{tabular} & \begin{tabular}[c]{@{}c@{}}\textbf{Milrinone}\\ (vasodilator)\end{tabular} \\
\begin{tabular}[c]{@{}c@{}}\textbf{\underline{Tirofiban}}\\ (prevention of blood clotting)\end{tabular} & \begin{tabular}[c]{@{}c@{}}\textbf{Pipazethate}\\ (antitussive)\end{tabular} & \begin{tabular}[c]{@{}c@{}}\textbf{Trapidil}\\ (vasodilator and anti-platelet agent)\end{tabular} \\ \hline
\begin{tabular}[c]{@{}c@{}}\textbf{\underline{Indapamide}}\\ (thiazide-like diuretic)\end{tabular} & \begin{tabular}[c]{@{}c@{}}\textbf{Mefenamic acid}\\ (anti-inflammatory)\end{tabular} & \begin{tabular}[c]{@{}c@{}}\textbf{Hydrochlorothiazide}\\ (thiazide diuretic)\end{tabular} \\
 decreases the metabolism & \begin{tabular}[c]{@{}c@{}}\textbf{Rolapitant}\\ (Neurokinin-1 receptor antagonist)\end{tabular} & \begin{tabular}[c]{@{}c@{}}\textbf{Chlorthalidone}\\ (thiazide-like diuretic)\end{tabular} \\
\begin{tabular}[c]{@{}c@{}}\textbf{Saquinavir}\\ (HIV protease inhibitor)\end{tabular} & \begin{tabular}[c]{@{}c@{}}\textbf{Kappadione}\\ (Vitamin K derivative)\end{tabular} & \begin{tabular}[c]{@{}c@{}}\textbf{Chlorothiazide}\\ (thiazide diuretic)\end{tabular} \\ \hline
\end{tabular}}
\label{table4}
\end{table*}
In the following we analyze and discuss the results observed in the above experiments and in order to demonstrate the effect of our model more graphically, include a visual representation of the model.

\subsection{Link Prediction}
\label{section5.1}
On both standard benchmark datasets, the knowledge graph embedding models using our adversarial framework gain better performance than existing methods on all metrics. KBGAN and RotatE also obtain good performance by introducing adversarial mechanisms to generate negative samples. The results demonstrate that the adversarial learning has the capability to construct more plausible triplets than random sampling does, and that these samples are conducive to improving the performance of the embedding models. The concrete adversarial mechanism proposed in this work differs from what is described in existing literature.

RotatE proposed a self-adversarial negative sampling mode, which selects negative triplets in accordance with scores calculated by the current KG embedding model. More Specifically, in the traditional model, the weight of each negative sample in the loss function is equally large:
\begin{equation}
    L=-\log \sigma\left(\gamma-f_{r}(\bm{h}, \bm{t})\right)-\sum_{i=1}^{n} \frac{1}{k} \log \sigma\left(f_{r}\left(\bm{h}_{i}^{\prime}, \bm{t}_{i}^{\prime}\right)-\gamma\right),
\end{equation}
where $\sigma$ denotes the sigmoid function, $\gamma$ indicates a fixed margin, and $f_{r}(\bm{h}_{i}^{\prime}, \bm{t}_{i}^{\prime})$ is the score of the $i$-th negative sample. RotatE first calculates the score of each negative sample according to the current embedding model, and then utilizes a Softmax operation to convert these scores into the weight corresponding to the negative sample. Finally, these weights are employed to calculate the total loss. Its mathematical definition is given by:
\begin{equation}
    p\left(h_{j}^{\prime}, r, t_{j}^{\prime} |\left\{\left(h_{i}, r_{i}, t_{i}\right)\right\}\right)=\frac{\exp \alpha f_{r}\left(\bm{h}_{j}^{\prime}, \bm{t}_{j}^{\prime}\right)}{\sum_{i} \exp \alpha f_{r}\left(\bm{h}_{i}^{\prime}, \bm{t}_{i}^{\prime}\right)},
\end{equation}
\begin{equation}
    L=-\log \sigma\left(\gamma-f_{r}(\bm{h}, \bm{t})\right)-\sum_{i=1}^{n} p\left(h_{i}^{\prime}, r, t_{i}^{\prime}\right) \log \sigma\left(f_{r}\left(\bm{h}_{i}^{\prime}, \bm{t}_{i}^{\prime}\right)-\gamma\right),
\end{equation}
where $\alpha$ indicates the temperature of sampling. Although the training time of this method is shorter than the general generative adversarial framework on account of removing the generator, the performance of the original RotatE is not as convincing as ours. As this original formulation lacks the game-style training process, the discriminator cannot compete with the generator and improve one another.

Our negative sampling strategy introduces an integral generative adversarial framework for better training of the discriminator that is similar to KBGAN. Compared with KBGAN, our proposed method has two advantages: 1) On top of the generator, we add a decoder module that turns the original negative sampling  into an autoencoder framework. This scheme ensures the fake drug generated by the generator to not deviate too far from the real one, thereby improving the plausibility of the negative sampling; 2) As it is not possible to use discrete data directly in the original GANs, where the discrete sampling step prevents gradients from propagating back to the generator, KBGAN relies on reinforcement learning to achieve its goal. However, increased computational cost and unstable training are inherent problems in reinforcement learning. Our method can solve both problems, as shown in Figure~\ref{figure4}.

To compare the negative samples generated by different sampling strategy more intuitively, we also visualize the corrupted triplets which are constructed by the generator and random sampling, respectively to further stress this point in Table~\ref{table4}.

Finally, it should be noted that this article focuses on comparing the impact of different negative sampling methods on model performance under the same conditions. As a consequence, we only utilize ranking-based loss functions which have a single output to construct our experiments. Instead, to obtain better end-to-end performance, knowledge graph embedding could also have been cast as a multi-class evaluation problem by employing multi-class based loss formulations that may lead to better downstream performance.

\subsection{DDI Classification}
\label{section5.2}
The same algorithm has different performance in link prediction and DDI classification tasks. This indicates that the two tasks measure different performance aspects of the knowledge graph embedding model. The result stresses the flexible adaptability and extensibility of our framework to different tasks.

The performance of RotatE varies greatly between the two datasets. A likely explanation lies in RotatE's fixed composition method~\cite{sun2018rotate}, utilizing the element-wise Hadamard product ($r_{1} \circ r_{2}$). For instance, given data on three persons $(a, b, c)$, where $b$ is the elder brother (marked as $r_{1}$) of $a$ and $c$ is the older sister (marked as $r_{2}$) of $b$, we can easily infer that $c$ is the older sister of $a$. The relation between $c$ and $a$ is $r_{2}$ rather than $r_{1} \circ r_{2}$. Perhaps this composition method can infer additional information when the number of available relations is small, but once the number of observed relations increases, this capability will no longer yield added value.

\subsection{Analysis of learned embeddings}
\label{section5.3}
\begin{figure*}
	\subfigure[]{
		\begin{minipage}[htb]{0.45\linewidth}
			\centering
			\includegraphics[height=2.2in]{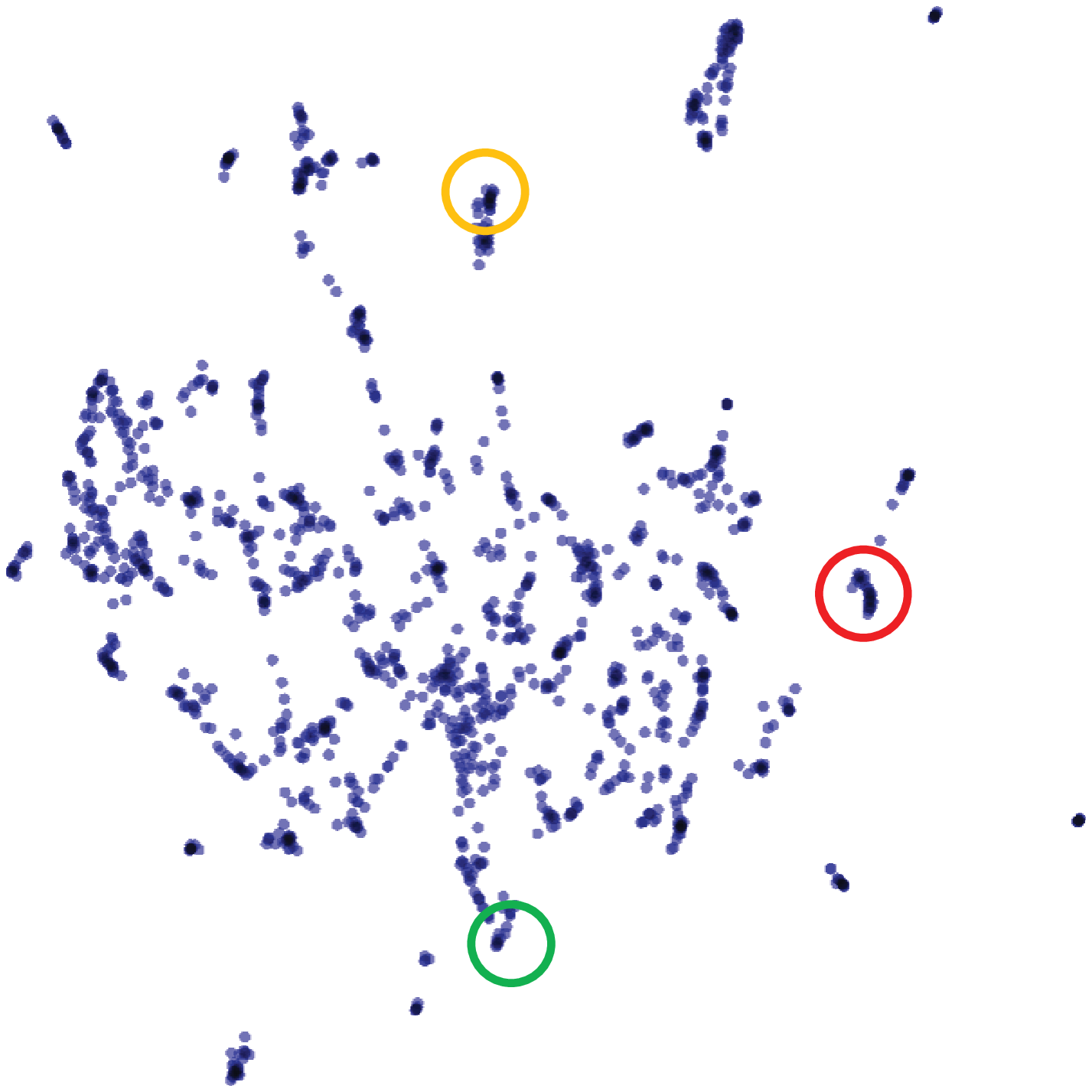}
			\label{figure3.1}
		\end{minipage}
	}
	\subfigure[]{
		\begin{minipage}[htb]{0.52\linewidth}
			\centering
			\includegraphics[height=2.2in]{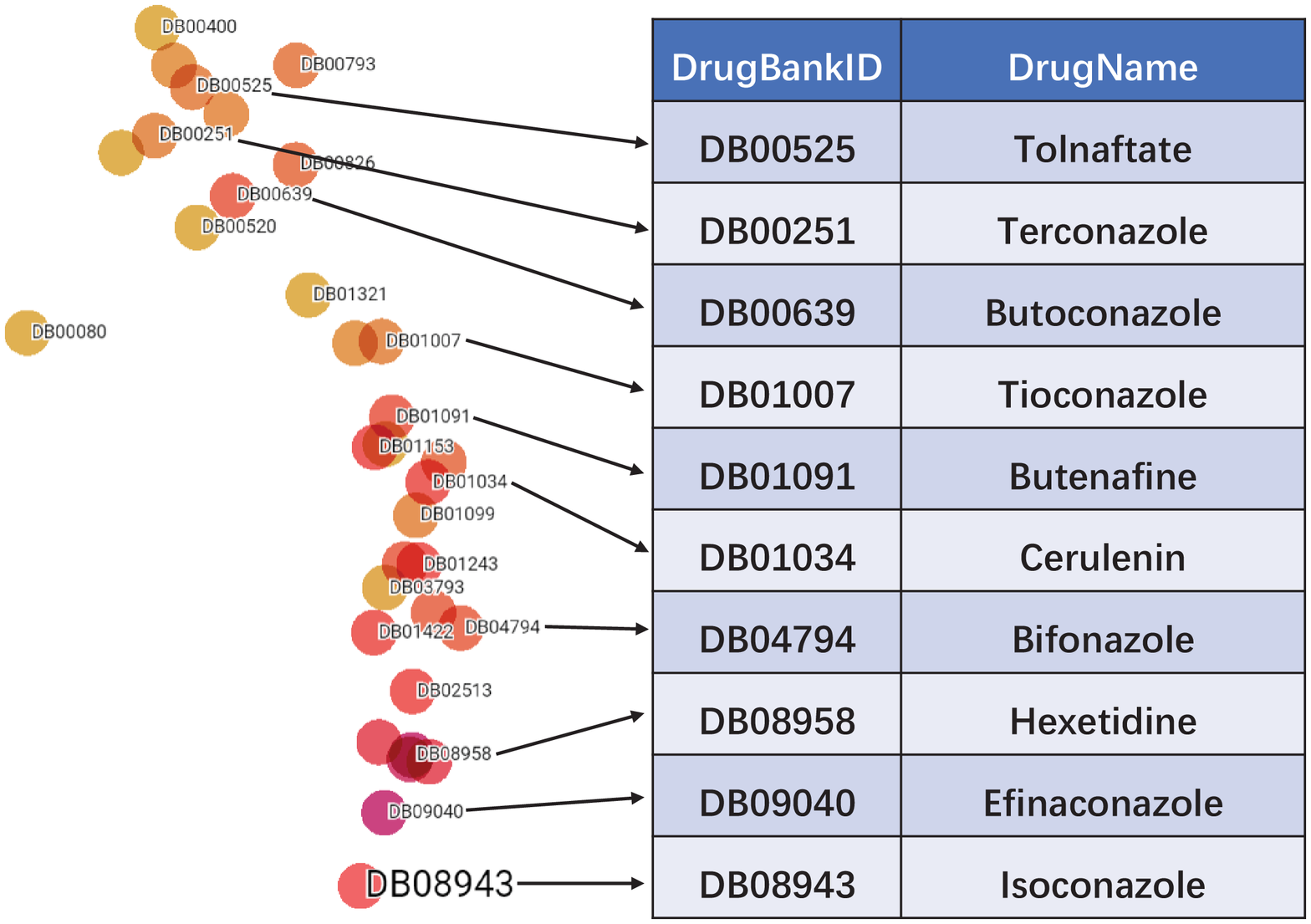}
			\label{figure3.2}
		\end{minipage}
	}
	\subfigure[]{
		\begin{minipage}[htb]{0.48\linewidth}
			\centering
			\includegraphics[height=2.13in]{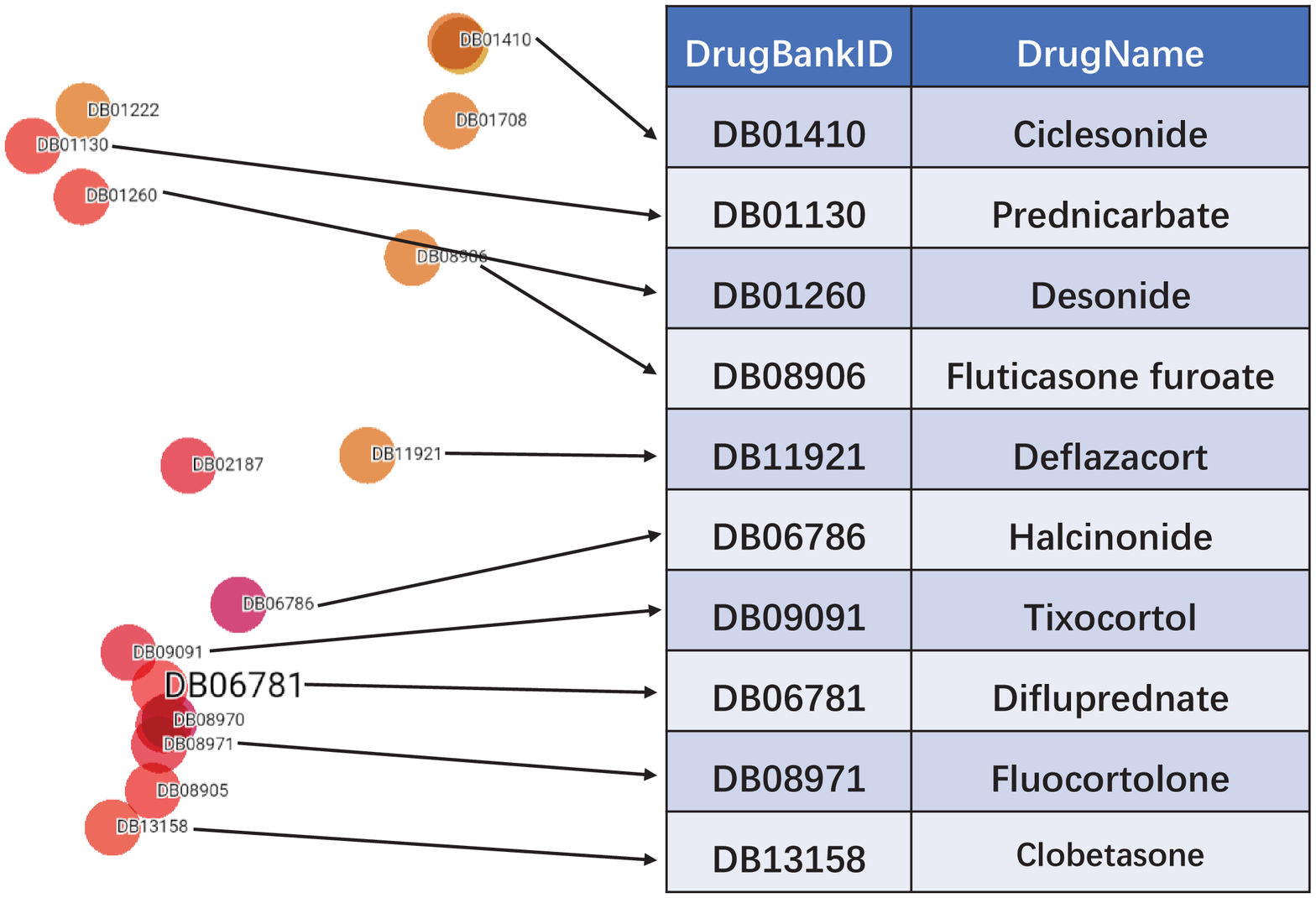}
			\label{figure3.3}
		\end{minipage}
	}
	\subfigure[]{
		\begin{minipage}[htb]{0.48\linewidth}
			\centering
			\includegraphics[height=2.2in]{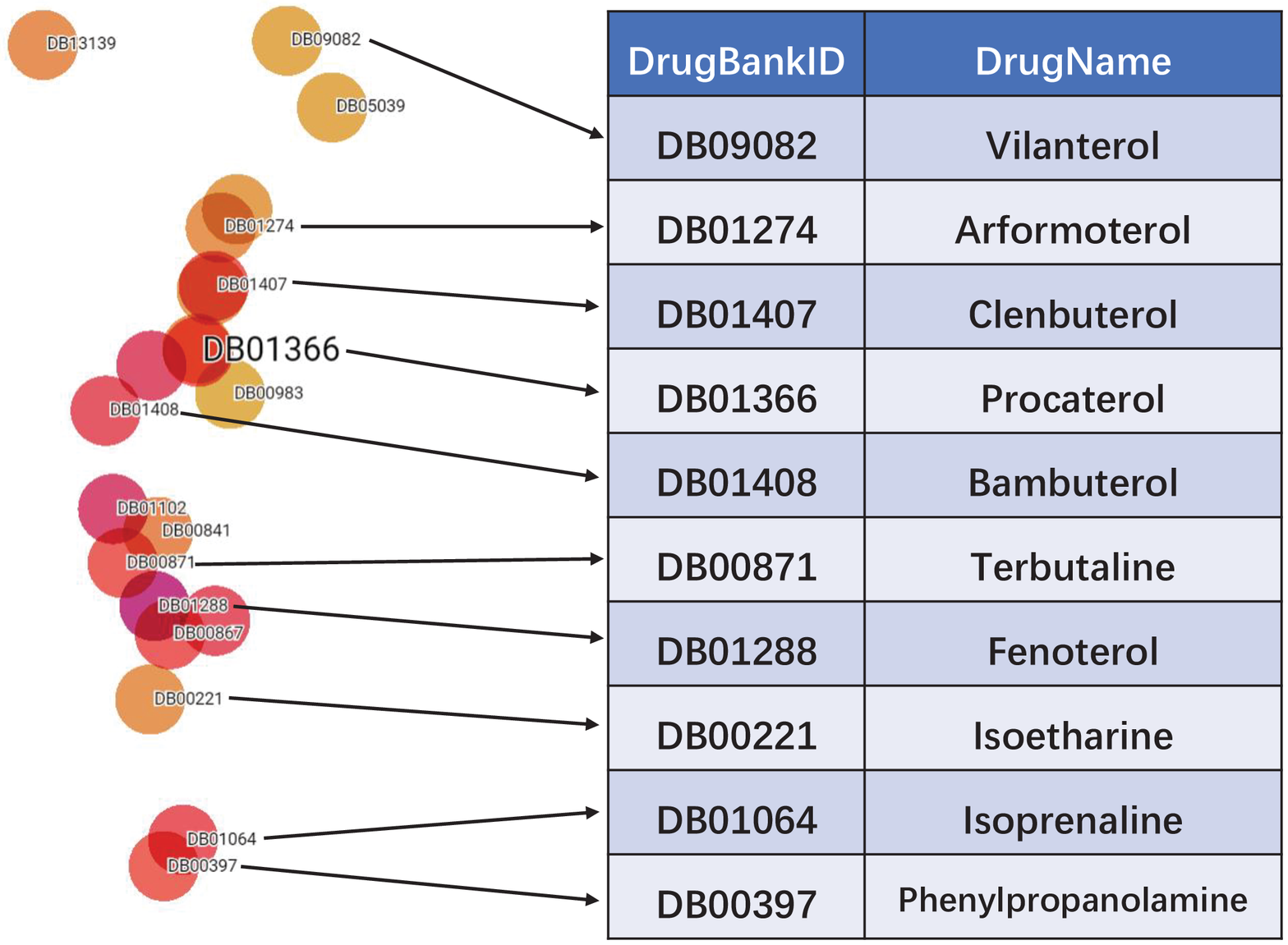}
			\label{figure3.4}
		\end{minipage}
	}
\caption{Illustrations of knowledge graph embedding vectors. (a) An overview of embedding vectors after dimensionality reduction, (b) The enlarged selection in the \textcolor[rgb]{1,0,0}{red} circle (the right one in (a)) contains mostly anti-fungal drugs. (c) The enlarged selection in the \textcolor[rgb]{0,1,0}{green} circle (the bottom one in (a)) contains mostly corticosteroids. (d) The enlarged selection in the \textcolor{yellow}{yellow} circle (the top one in (a)) contains asthma medication. As we can see, the drugs in each of the above circles have similar effects or categories. This illustration proves that our model follows the translation invariance criterion.}
\label{fig3}
\end{figure*}
To highlight the capabilities of our proposed framework in a qualitative manner we include two visualization experiments: negative samples constructed by random mode versus the generator, and an illustration of knowledge graph embedding vectors. It is worth to note that our model has access to only the structure of the graph in the form of triplets and is agnostic of any other features and properties. Therefore, we cannot easily utilize visualization via attention mechanisms, as our input does not represent specific features or characteristics.

Table~\ref{table4}, compares traditional random negative samples with generated ones. We can note that the generator is able to select more semantically relevant drugs as negative samples. For instance, given a real triplet (\emph{Midazolam, increases the risk of adverse effects, Dezocine}), the generator adopts three tail drugs, \textit{i.e.} \emph{Methadyl acetate}, \emph{Levacetylmethadol} and \emph{Nalbuphine}. All three drugs, similar to \emph{Dezocine}, have narcotic effects. Thus, the negative triplets constructed by these drugs are more plausible and potentially deceptive.

Given such high-quality negative triplets, we can train better knowledge graph embedding models which have strengthened representation and generalization capabilities. Similar to word embedding, knowledge graph embedding also follows a basic principle that entities with similar connotation should have similar representations. We demonstrate this by projecting the trained drug vectors into a two-dimensional space to validate whether they satisfy this principle. Figure~\ref{fig3} shows an illustration of KG embedding vectors after dimensionality reduction.

Embedding vectors are projected to two-dimensional space by applying UMAP dimensionality reduction~\cite{mcinnes2018umap} (Figure \ref{figure3.1}). We select three regions of the embedding space and zoom in on them to observe if the drugs in those areas are related in indication or category. Figure~\ref{figure3.2} lists 10 drugs in the red circle with consistently anti-fungal effect. Similarly, the vast majority of drugs in the yellow region are corticosteroids, and the green region contains asthma medication. This illustration also intuitively supports the effectiveness of our model from a qualitative perspective.

It should be noted that this article focuses on improving the performance of KG embedding models, while ignoring the comparison with clinical trials or traditional machine learning based methods. These older methods often do not provide datasets or the datasets are too small to train modern KGE models such as ours. We hope to remedy this limitation in future work.

\subsection{Complexity and training time analysis}
\label{section5.4}
The training time complexity of all models introduced in this experiment, including TransE, DistMult, ComplEx, SimplE and RotatE, is $\mathcal{O}(d)$ where $d$ denotes the dimensionality of the embedding space. Since KBGAN and our proposed framework both consist of two parts, a generator (an autoencoder for our framework) with $\mathcal{O}(d)$ time complexity and a discriminator with $\mathcal{O}(d)$ time complexity, the time complexity of these two frameworks remains linear in $d$. However, as we mentioned above, policy gradient based methods can be unstable in the training process, while our proposed method can complete optimization more efficiently.

\begin{figure*}[]
	\begin{minipage}{0.32\linewidth}
		\centerline{\includegraphics[height=3.5cm]{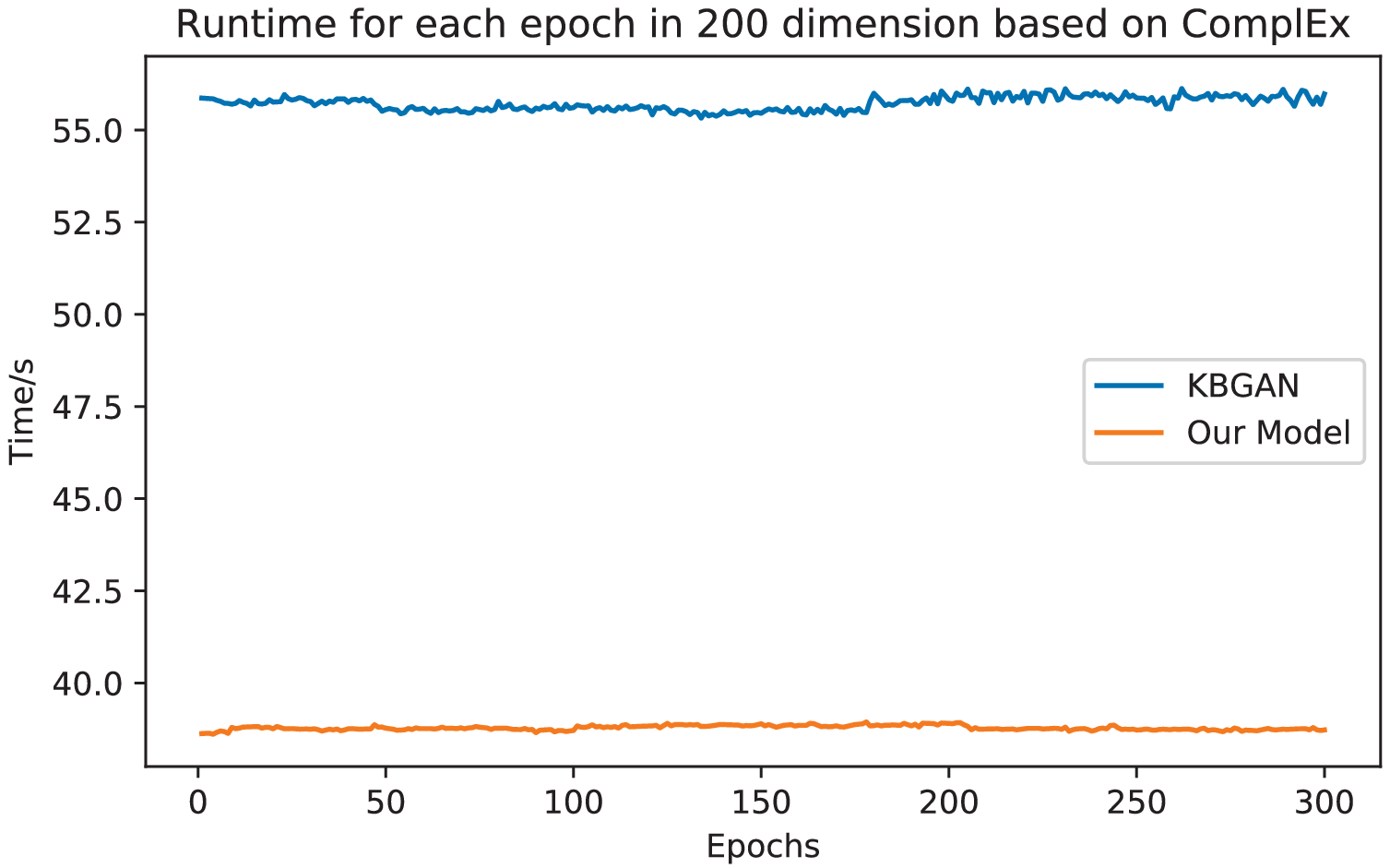}}
	\end{minipage}
	%\hfill
	\begin{minipage}{0.32\linewidth}
		\centerline{\includegraphics[height=3.5cm]{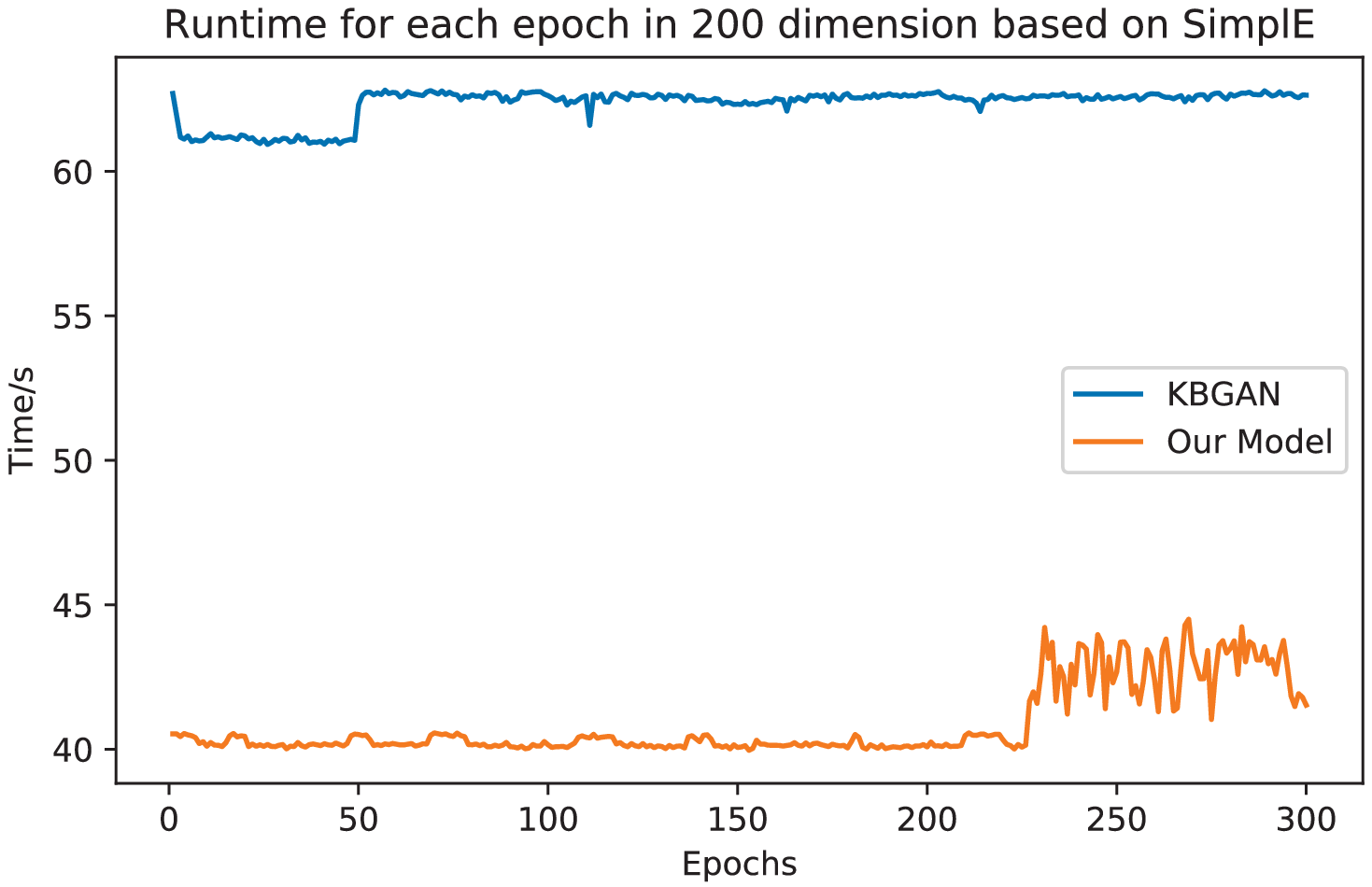}}
		%  \centerline{\input{fig5b.pdf_tex}}
	\end{minipage}
	%\vfill
	\begin{minipage}{0.32\linewidth}
		\centerline{\includegraphics[height=3.5cm]{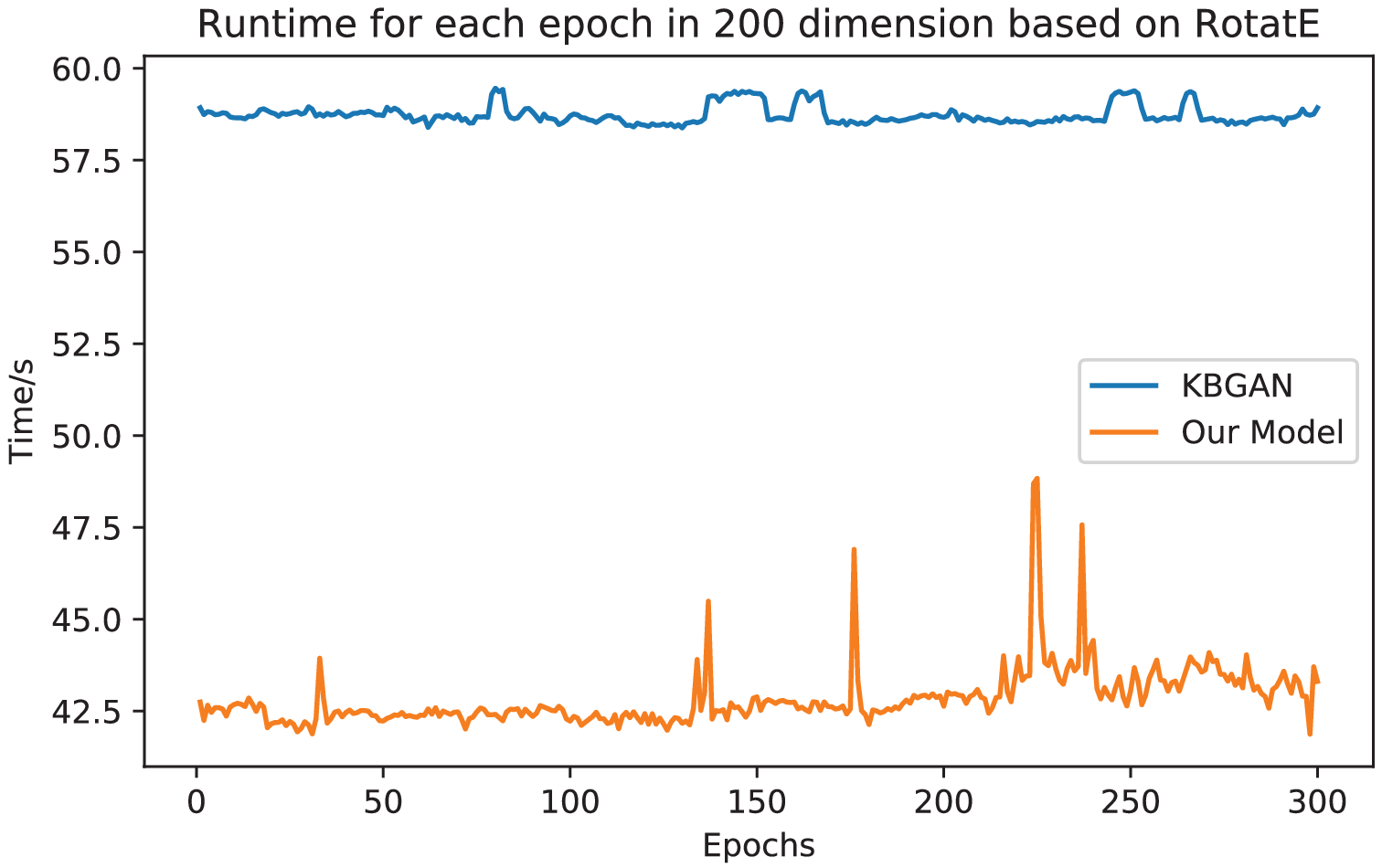}}
		%  \centerline{\input{fig5c.pdf_tex}}
	\end{minipage}
	%\hfill
	\setlength{\belowcaptionskip}{-0.4cm}
	%\vspace{-2mm}
	\caption{Runtimes for each epoch on DeepDDI dataset. We plot the training times of KBGAN and our model in which ComplEx, SimplE and RotatE are as their discriminator in 200 dimension respectively.}
\label{figure4}
	%\vspace{-1mm}
\end{figure*}

Figure~\ref{figure4} plots the runtime of 300 training epochs of KBGAN and our model on DeepDDI. Both models use ComplEx, SimplE and RotatE as their discriminator. The result indicates that our method effectively shortens the training time compared with policy gradient based GANs. Moreover, because our model and KBGAN both can be divided into two major modules: generator and discriminator, the number of parameters will be greater than that of a single embedding model. Thus, although their formal complexity is the same as that of single embedding models, these two adversarial learning frameworks will require more time to complete the training process.

In addition to the complexity of the model, the size of the knowledge graph, including the total number of entities and relations and the resulting number of triplets, are also significant indicators that affect training time. Models on the DeepDDI dataset require more time to train than those on Decagon, because the number of entities in DeepDDI is significantly greater than in Decagon.

%%%%%%%%%%%%%%%%%%%%%%%%%%%%%%%%%%%%%%%%%%%%%%%%%%%%%%%%%%%%%%%%%%%%%%%%%%%%%%%%%
\section{Conclusions}
\label{section6}
The goal of this study is to find a new approach to negative sampling that improves the performance of drug-drug interaction knowledge graph embedding models. In this paper, we propose an adversarial learning framework based on Wasserstein distances for this task. We evaluate the proposed method on link prediction and DDI classification tasks. Our experiments on two standard collections confirm that the performance of all baseline models can be significantly improved using our adversarial learning framework. 

The approach has several major advantages over existing knowledge graph embedding models. First, we introduce an adversarial autoencoder framework to represent drug-drug interaction knowledge graphs. The autoencoder is employed to generate more plausible drugs as negative samples, and these negative triplets are fed to the discriminator along with authentic positive ones for improving the performance of embedding models. Our approach also utilizes a Gumbel-Softmax relaxation and Wasserstein distance to handle vanishing gradient issues on discrete data. Compared with traditional policy gradients in reinforcement learning, the proposed method can complete optimization tasks more efficiently. Most notably, the work presented here can be applied to refine the performance of most existing models without the need for major modifications.

Our method is not limited to the DDI domain. Going beyond the application and scope of this immediate work, future work will include evaluating the benefits the model presented here holds for other graph embedding tasks such as recommendation, classification and retrieval settings on hierarchical data. 
% \textcolor[rgb]{0,0,1}{The last thing required a bit of explanation is that this article focuses on comparing the performance of different KG embedding models, while ignoring the comparison with some traditional clinical trials or machine learning based methods, because these traditional methods are relatively old, most of the mentioned datasets are missing or too small to train our proposed model. We hope to find some suitable datasets to remedy for this limitation in the future work.}

\section*{Acknowledgement}
This work is supported by the China Scholarship Council (CSC) Grant No. 201906650004 and the National Natural Science Foundation of China (Nos. U1705262 and 61672159).
\bibliographystyle{unsrt}
\bibliography{mybibfile.bib}
\end{document}